\documentclass[12pt]{amsart}

\usepackage{amsmath}
\usepackage{amssymb}
\usepackage{amsfonts}
\usepackage{latexsym}
\usepackage{color}
\usepackage{graphicx}
\usepackage{natbib}
\usepackage{times}
\usepackage{fixmath}

\newcommand{\citeN}{\cite}

\newcommand{\Dir}{\textrm{Dirichlet}}

\newenvironment{packed_enumerate}{
  \begin{enumerate}
    \setlength{\topsep}{0pt}
    \setlength{\itemsep}{2pt}
    \setlength{\parskip}{0pt}
    \setlength{\parsep}{0pt}}
  {\end{enumerate}}

\newcommand{\cal}{\mathcal}
\newcommand{\bet}{\textrm{Beta}}

\newcommand{\cc}{\textbf{c}}
\newcommand{\g}{\,\vert\,}
\newcommand{\mult}{\textrm{Mult}}
\newcommand{\dir}{\textrm{Dir}}
\newcommand{\E}{\textrm{E}}

\newcommand{\mysec}[1]{Section~\ref{sec:#1}}

\newcommand{\myeq}[1]{Eq.~(\ref{eq:#1})}

\newcommand{\myfig}[1]{Figure~\ref{fig:#1}}
\newcommand{\vct}[1]{\textbf{#1}}
\newcommand{\vctg}[1]{\mathbold{#1}}

\newcommand{\cnt}[1]{\#[#1]}
\newcommand{\gem}{\textrm{GEM}}

\newcommand{\BibTeX}{{\rm B\kern-.05em{\sc i\kern-.025em b}\kern-.08em
    T\kern-.1667em\lower.7ex\hbox{E}\kern-.125emX}}

\begin{document}

\title[The nested Chinese restaurant process]{The nested Chinese
  restaurant process and Bayesian nonparametric inference of topic
  hierarchies}

\author[D. M. Blei, T. L. Griffiths, M. I. Jordan]
{David M. Blei \\ \textit{Princeton University}
  \\ \\
  Thomas L. Griffiths \\ \textit{University of California, Berkeley}
  \\ \\
  Michael I. Jordan \\ \textit{University of California, Berkeley}
}

\begin{abstract}
  We present the nested Chinese restaurant process (nCRP), a
  stochastic process which assigns probability distributions to
  infinitely-deep, infinitely-branching trees.  We show how this
  stochastic process can be used as a prior distribution in a Bayesian
  nonparametric model of document collections.  Specifically, we
  present an application to information retrieval in which documents
  are modeled as paths down a random tree, and the preferential
  attachment dynamics of the nCRP leads to clustering of documents
  according to sharing of topics at multiple levels of abstraction.
  Given a corpus of documents, a posterior inference algorithm finds
  an approximation to a posterior distribution over trees, topics and
  allocations of words to levels of the tree.  We demonstrate this
  algorithm on collections of scientific abstracts from several
  journals.  This model exemplifies a recent trend in statistical
  machine learning---the use of Bayesian nonparametric methods to
  infer distributions on flexible data structures.
\end{abstract}

\maketitle

\section{Introduction}

For much of its history, computer science has focused on deductive
formal methods, allying itself with deductive traditions in areas of
mathematics such as set theory, logic, algebra, and combinatorics.
There has been accordingly less focus on efforts to develop inductive,
empirically-based formalisms in computer science, a gap which became
increasingly visible over the years as computers have been required to
interact with noisy, difficult-to-characterize sources of data, such
as those deriving from physical signals or from human activity.  In
more recent history, the field of machine learning has aimed to fill
this gap, allying itself with inductive traditions in probability and
statistics, while focusing on methods that are amenable to analysis as
computational procedures.

% !!! dmb: above, i think we should include a footnote about the
% probabilistic method and its different way of using probability.
% MJ: Not sure I agree; the probabilistic method is a purely deductive
% idea, and our point (about induction in CS) stands.  Note by the way
% that the probabilistic method wasn't developed in CS per se; it was
% a standard idea in probability theory since Kolmogorov, and its
% flowering outside of probability came in combinatorics first, then
% in theoretical CS.

% Loosely, machine learning methods can be divided into two groups:
% supervised methods and unsupervised methods.  MJ: I removed this
% sentence, because a lot of ML people work in reinforcement learning,
% which isn't in one of these two groups.  This required rewriting the
% paragraph.

Machine learning methods can be divided into \emph{supervised
  learning} methods and \emph{unsupervised learning} methods.
Supervised learning has been a major focus of machine learning
research.  In supervised learning, each data point is associated with
a label (e.g., a category, a rank or a real number) and the goal is to
find a function that maps data into labels (so as to predict the
labels of data that have not yet been labeled).  A canonical example
of supervised machine learning is the email spam filter, which is
trained on known spam messages and then used to mark incoming
unlabeled email as spam or non-spam.

While supervised learning remains an active and vibrant area of
research, more recently the focus in machine learning has turned to
unsupervised learning methods.  In unsupervised learning the data are
not labeled, and the broad goal is to find patterns and structure
within the data set.  Different formulations of unsupervised learning
are based on different notions of ``pattern'' and ``structure.''
Canonical examples include \emph{clustering}, the problem of grouping
data into meaningful groups of similar points, and \emph{dimension
  reduction}, the problem of finding a compact representation that
retains useful information in the data set.  One way to render these
notions concrete is to tie them to a supervised learning problem;
thus, a structure is validated if it aids the performance of an
associated supervised learning system.  Often, however, the goal is
more exploratory.  Inferred structures and patterns might be used, for
example, to visualize or organize the data according to subjective
criteria.  With the increased access to all kinds of unlabeled
data---scientific data, personal data, consumer data, economic data,
government data, text data---exploratory unsupervised machine learning
methods have become increasingly prominent.

Another important dichotomy in machine learning distinguishes between
\emph{parametric} and \emph{nonparametric} models.  A parametric model
involves a fixed representation that does not grow structurally as
more data are observed.  Examples include linear regression and
clustering methods in which the number of clusters is fixed a priori.
A nonparametric model, on the other hand, is based on representations
that are allowed to grow structurally as more data are
observed.\footnote{In particular, despite the nomenclature, a
  nonparametric model can involve parameters; the issue is whether or
  not the number of parameters grows as more data are observed.}
Nonparametric approaches are often adopted when the goal is to impose
as few assumptions as possible and to ``let the data speak.''

The nonparametric approach underlies many of the most significant
developments in the supervised learning branch of machine learning
over the past two decades.  In particular, modern classifiers such as
decision trees, boosting and nearest neighbor methods are
nonparametric, as are the class of supervised learning systems built
on ``kernel methods,'' including the support vector machine.
(See~\cite{Hastie:2001} for a good review of these methods.)
Theoretical developments in supervised learning have shown that as the
number of data points grows, these methods can converge to the true
labeling function underlying the data, even when the data lie in an
uncountably infinite space and the labeling function is
arbitrary~\citep{Devroye96}.  This would clearly not be possible for
parametric classifiers.

The assumption that labels are available in supervised learning is a
strong assumption, but it has the virtue that few additional
assumptions are generally needed to obtain a useful supervised
learning methodology.  In unsupervised learning, on the other hand,
the absence of labels and the need to obtain operational definitions
of ``pattern'' and ``structure'' generally makes it necessary to
impose additional assumptions on the data source.  In particular,
unsupervised learning methods are often based on ``generative
models,'' which are probabilistic models that express hypotheses about
the way in which the data may have been generated.
\emph{Probabilistic graphical models} (also known as ``Bayesian
networks'' and ``Markov random fields'') have emerged as a broadly
useful approach to specifying generative
models~\citep{Lauritzen:1996,Jordan:2000}.  The elegant marriage of
graph theory and probability theory in graphical models makes it
possible to take a fully probabilistic (i.e., Bayesian) approach to
unsupervised learning in which efficient algorithms are available to
update a prior generative model into a posterior generative model once
data have been observed.

Although graphical models have catalyzed much research in unsupervised
learning and have had many practical successes, it is important to
note that most of the graphical model literature has been focused on
parametric models.  In particular, the graphs and the local potential
functions comprising a graphical model are viewed as fixed objects;
they do not grow structurally as more data are observed.  Thus, while
nonparametric methods have dominated the literature in supervised
learning, parametric methods have dominated in unsupervised learning.
This may seem surprising given that the open-ended nature of the
unsupervised learning problem seems particularly commensurate with the
nonparametric philosophy.  But it reflects an underlying tension in
unsupervised learning---to obtain a well-posed learning problem it is
necessary to impose assumptions, but the assumptions should not be too
strong or they will inform the discovered structure more than the data
themselves.

It is our view that the framework of \emph{Bayesian nonparametric
  statistics} provides a general way to lessen this tension and to
pave the way to unsupervised learning methods that combine the virtues
of the probabilistic approach embodied in graphical models with the
nonparametric spirit of supervised learning.  In Bayesian
nonparametric (BNP) inference, the prior and posterior distributions
are no longer restricted to be parametric distributions, but are
general \emph{stochastic processes}~\cite{HjortEtAl}.  Recall that a
stochastic process is simply an indexed collection of random
variables, where the index set is allowed to be infinite.  Thus, using
stochastic processes, the objects of Bayesian inference are no longer
restricted to finite-dimensional spaces, but are allowed to range over
general infinite-dimensional spaces.  For example, objects such as
trees of arbitrary branching factor and arbitrary depth are allowed
within the BNP framework, as are other structured objects of
open-ended cardinality such as partitions and lists.  It is also
possible to work with stochastic processes that place distributions on
functions and distributions on distributions.  The latter fact
exhibits the potential for recursive constructions that is available
within the BNP framework.  In general, we view the representational
flexibility of the BNP framework as a statistical counterpart of the
flexible data structures that are ubiquitous in computer science.

In this paper, we aim to introduce the BNP framework to a wider
computational audience by showing how BNP methods can be deployed in a
specific unsupervised machine learning problem of significant current
interest---that of learning \emph{topic models} for collections of
text, images and other semi-structured
corpora~\cite{Blei:2003b,Griffiths:2006,Blei:2009}.

Let us briefly introduce the problem here; a more formal presentation
appears in \mysec{hlda}.  A \emph{topic} is defined to be a
probability distribution across \emph{words} from a \emph{vocabulary}.
Given an input \emph{corpus}---a set of \emph{documents} each
consisting of a sequence of words---we want an algorithm to both find
useful sets of topics and learn to organize the topics according to a
hierarchy in which more abstract topics are near the root of the
hierarchy and more concrete topics are near the leaves.  While a
classical unsupervised analysis might require the topology of the
hierarchy (branching factors, etc) to be chosen in advance, our BNP
approach aims to infer a distribution on topologies, in particular
placing high probability on those hierarchies that best explain the
data.  Moreover, in accordance with our goals of using flexible models
that ``let the data speak,'' we wish to allow this distribution to
have its support on arbitrary topologies---there should be no
limitations such as a maximum depth or maximum branching factor.

\begin{figure}
  \begin{center}
    \includegraphics[width=\textwidth]
    {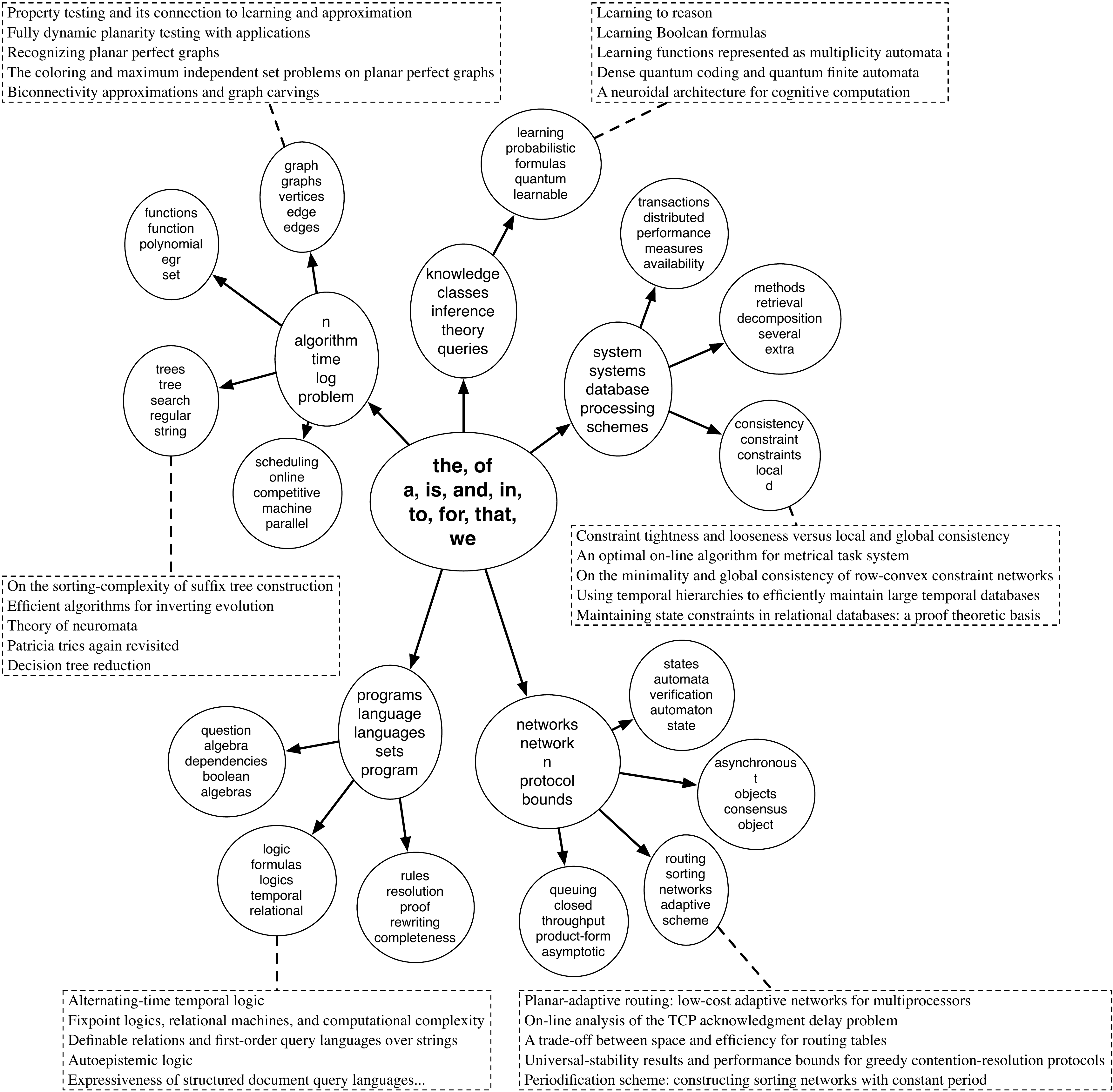}
  \end{center}
  \caption{\label{fig:jacm} The topic hierarchy learned from 536
    abstracts of the \textit{Journal of the ACM} (JACM) from
    1987--2004.  The vocabulary was restricted to the 1,539 terms that
    occurred in more than five documents, yielding a corpus of 68K words.
    The learned hierarchy contains 25 topics, and each topic node is
    annotated with its top five most probable terms.  We also present
    examples of documents associated with a subset of the paths in the
    hierarchy.}
\end{figure}

We provide an example of the output from our algorithm in
\myfig{jacm}.
The input corpus in this case was a collection of abstracts from the
\textit{Journal of the ACM} (JACM) from the years 1987 to 2004.  The
figure depicts a topology that is given highest probability by our
algorithm, along with the highest probability words from the topics
associated with this topology (each node in the tree corresponds to a
single topic).  As can be seen from the figure, the algorithm has
discovered the category of function words at level zero (e.g., ``the''
and ``of''), and has discovered a set of first-level topics that are
a reasonably faithful representation of some of the main areas of
computer science.  The second level provides a further subdivision
into more concrete topics.  We emphasize that this is an unsupervised
problem.  The algorithm discovers the topic hierarchy without any
extra information about the corpus (e.g., keywords, titles or
authors).  The documents are the only inputs to the algorithm.

A learned topic hierarchy can be useful for many tasks, including text
categorization, text compression, text summarization and language
modeling for speech recognition.  A commonly-used surrogate for the
evaluation of performance in these tasks is predictive likelihood, and
we use predictive likelihood to evaluate our methods quantitatively.
But we also view our work as making a contribution to the development
of methods for the visualization and browsing of documents.  The model
and algorithm we describe can be used to build a topic hierarchy for a
document collection, and that hierarchy can be used to sharpen a
user's understanding of the contents of the collection.  A qualitative
measure of the success of our approach is that the same tool should be
able to uncover a useful topic hierarchy in different domains based
solely on the input data.

By defining a probabilistic model for documents, we do not define the
level of ``abstraction'' of a topic formally, but rather define a
statistical procedure that allows a system designer to capture notions
of abstraction that are reflected in usage patterns of the specific
corpus at hand.  While the content of topics will vary across corpora,
the ways in which abstraction interacts with usage will not. A corpus
might be a collection of images, a collection of HTML documents or a
collection of DNA sequences.  Different notions of abstraction will be
appropriate in these different domains, but each are expressed and
discoverable in the data, making it possible to automatically
construct a hierarchy of topics.

This paper is organized as follows.  We begin with a review of the
necessary background in stochastic processes and Bayesian
nonparametric statistics in \mysec{background}.  In \mysec{nCRP}, we
develop the nested Chinese restaurant process, the prior on topologies
that we use in the hierarchical topic model of \mysec{hlda}.  We
derive an approximate posterior inference algorithm in
\mysec{posterior} to learn topic hierarchies from text data.  Examples
and an empirical evaluation are provided in \mysec{results}.  Finally,
we present related work and a discussion in \mysec{discussion}.

\section{Background}
\label{sec:background}

Our approach to topic modeling reposes on several building blocks from
stochastic process theory and Bayesian nonparametric statistics,
specifically the Chinese restaurant process~\citep{Aldous:1985},
stick-breaking processes~\citep{Pitman:2002}, and the Dirichlet
process mixture~\citep{Antoniak:1974}.  In this section we briefly 
review these ideas and the connections between them.

\subsection{Dirichlet and beta distributions}

Recall that the \emph{Dirichlet distribution} is a probability distribution
on the simplex of nonnegative real numbers that sum to one.  We write
\[
U \sim \dir(\alpha_1, \alpha_2, \ldots, \alpha_K),
\]
for a random vector $U$ distributed as a Dirichlet random variable on
the $K$-simplex, where $\alpha_i > 0$ are parameters.  The mean of $U$
is proportional to the parameters
\[
E[U_i] = \frac{\alpha_i}{\sum_{k=1}^K \alpha_k}
\]
and the magnitude of the parameters determines the concentration of $U$
around the mean.  The specific choice $\alpha_1 = \cdots = \alpha_K = 1$
yields the uniform distribution on the simplex.  Letting $\alpha_i > 1$
yields a unimodal distribution peaked around the mean, and letting
$\alpha_i < 1$ yields a distribution that has modes at the corners of the
simplex. The \emph{beta distribution} is a special case of the Dirichlet
distribution for $K=2$, in which case the simplex is the unit interval
$(0,1)$.  In this case we write $U \sim \bet(\alpha_1, \alpha_2)$, where
$U$ is a scalar.

\subsection{Chinese restaurant process}
\label{sec:CRP}

\begin{figure}
  \begin{center}
    \includegraphics[width=0.9 \textwidth]{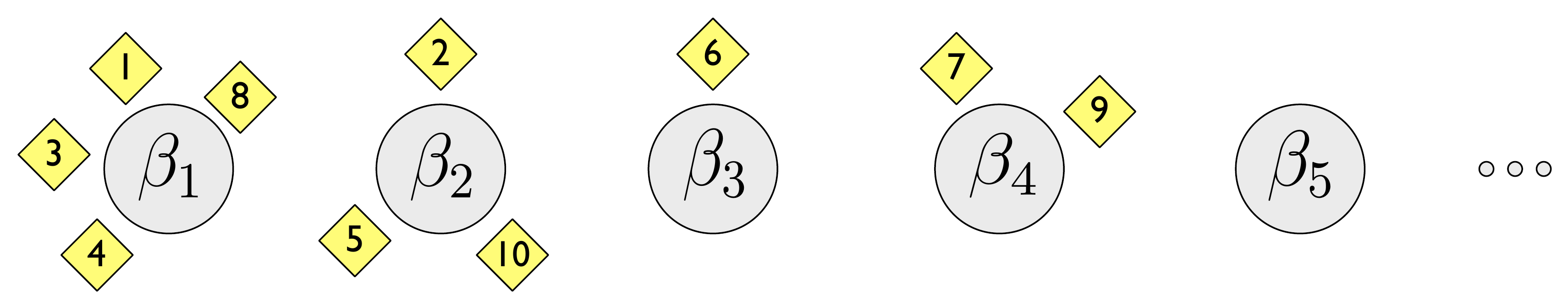}
  \end{center}
  \caption{\label{fig:CRP} A configuration of the Chinese restaurant
    process.  There are an infinite number of tables, each associated
    with a parameter $\beta_i$.  The customers sit at the tables
    according to \myeq{CRP} and each generate data with the
    corresponding parameter.  In this configuration, ten customers
    have been seated in the restaurant, populating four of the
    infinite set of tables.}
\end{figure}

The Chinese restaurant process (CRP) is a single parameter
distribution over partitions of the integers.  The distribution can be
most easily described by specifying how to draw a sample from it.
Consider a restaurant with an infinite number of tables each with
infinite capacity.  A sequence of $N$ customers arrive, labeled with
the integers $\{1, \ldots, N\}$.  The first customer sits at the first
table; the $n$th subsequent customer sits at a table drawn from the
following distribution:
\begin{equation}
  \label{eq:CRP}
  \begin{array}{rcl}
    p(\textrm{occupied table $i$} \g \textrm{previous customers}) &=&
    \frac{n_{i}}{\gamma + n - 1} \\
    p(\textrm{next unoccupied table} \g \textrm{previous customers}) &=&
    \frac{\gamma}{\gamma + n - 1},
  \end{array}
\end{equation}
where $n_i$ is the number of customers currently sitting at table $i$,
and $\gamma$ is a real-valued parameter which controls how often,
relative to the number of customers in the restaurant, a customer
chooses a new table versus sitting with others.  After $N$ customers
have been seated, the seating plan gives a partition of those
customers as illustrated in \myfig{CRP}.

With an eye towards Bayesian statistical applications, we assume that
each table is endowed with a parameter vector $\beta$ drawn from a
distribution $G_0$.  Each customer is associated with the parameter
vector at the table at which he sits.  The resulting distribution on
sequences of parameter values is referred to as a \emph{P\'olya urn
  model}~\citep{Johnson:1977}.

The P\'olya urn distribution can be used to define a flexible
clustering model.  Let the parameters at the tables index a family of
probability distributions (for example, the distribution might be a
multivariate Gaussian in which case the parameter would be a mean
vector and covariance matrix).  Associate customers to data points,
and draw each data point from the probability distribution 
associated with the table at which the customer sits.  This 
induces a probabilistic clustering of the generated data because 
customers sitting around each table share the same parameter vector.

This model is in the spirit of a traditional mixture
model~\citep{Titterington:1985}, but is critically different in that
the number of tables is unbounded.  Data analysis amounts to
\textit{inverting} the generative process to determine a probability
distribution on the ``seating assignment'' of a data set.  The
underlying CRP lets the data determine the number of clusters (i.e.,
the number of occupied tables) and further allows new data to be
assigned to new clusters (i.e., new tables).

\subsection{Stick-breaking constructions}
\label{sec:stick-breaking}

The Dirichlet distribution places a distribution on nonnegative
$K$-dimensional vectors whose components sum to one.  In this section
we discuss a stochastic process that allows $K$ to be unbounded.

Consider a collection of nonnegative real numbers
$\{\theta_i\}_{i=1}^\infty$ where $\sum_i \theta_i = 1$.  We wish to
place a probability distribution on such sequences.  Given that each
such sequence can be viewed as a probability distribution on the
positive integers, we obtain a distribution on distributions, i.e., a
random probability distribution.

To do this, we use a \emph{stick-breaking construction}.  View the
interval $(0,1)$ as a unit-length stick.  Draw a value $V_1$ from a
$\bet(\alpha_1, \alpha_2)$ distribution and break off a fraction $V_1$
of the stick.  Let $\theta_1 = V_1$ denote this first fragment of the
stick and let $1 - \theta_1$ denote the remainder of the stick.
Continue this procedure recursively, letting $\theta_2 = V_2 (1 -
\theta_1)$, and in general define
\[
\theta_i = V_i \prod_{j=1}^{i-1} (1 - V_j),
\]
where $\{V_i\}$ are an infinite sequence of independent draws from the
$\bet(\alpha_1, \alpha_2)$ distribution.  \citeN{Sethuraman:1994}
shows that the resulting sequence $\{\theta_i\}$ satisfies $\sum_i
\theta_i = 1$ with probability one.

In the special case $\alpha_1 = 1$ we obtain a one-parameter
stochastic process known as the \emph{GEM
  distribution}~\citep{Pitman:2002}.  Let $\gamma = \alpha_2$ denote
this parameter and denote draws from this distribution as $\theta \sim
\gem(\gamma)$.  Large values of $\gamma$ skew the beta distribution
towards zero and yield random sequences that are heavy-tailed, i.e.,
significant probability tends to be assigned to large integers.  Small
values of $\gamma$ yield random sequences that decay more quickly to
zero.

\subsection{Connections}
\label{sec:connections}

The GEM distribution and the CRP are closely related.  Let $\theta
\sim \gem(\gamma)$ and let $\{Z_1, Z_2, \ldots, Z_N\}$ be a sequence
of indicator variables drawn independently from $\theta$, i.e.,
\[
p(Z_n = i \g \theta) = \theta_i.
\]
This distribution on indicator variables induces a random partition on
the integers $\{1, 2, \ldots, N\}$, where the partition reflects
indicators that share the same values.  It can be shown that this
distribution on partitions is the same as the distribution on
partitions induced by the CRP~\citep{Pitman:2002}.  As implied by this
result, the GEM parameter $\gamma$ controls the partition in the same
way as the CRP parameter $\gamma$.

As with the CRP, we can augment the GEM distribution to consider draws
of parameter values.  Let $\{\beta_i\}$ be an infinite sequence of
independent draws from a distribution $G_0$ defined on a sample space
$\Omega$.  Define
\[
G = \sum_{i=1}^{\infty} \theta_i \delta_{\beta_i},
\]
where $\delta_{\beta_i}$ is an atom at location $\beta_i$ and where
$\theta \sim \gem(\gamma)$.  The object $G$ is a distribution on
$\Omega$; it is a \emph{random distribution}.

Consider now a finite partition of $\Omega$.  \citeN{Sethuraman:1994}
showed that the probability assigned by $G$ to the cells of this
partition follows a Dirichlet distribution.  Moreover, if we consider
all possible finite partitions of $\Omega$, the resulting Dirichlet
distributions are consistent with each other.  Thus, by appealing to
the Kolmogorov consistency theorem~\citep{Billingsley:1995}, we can
view $G$ as a draw from an underlying stochastic process, where the
index set is the set of Borel sets of $\Omega$.  This stochastic 
process is known as the \emph{Dirichlet process}~\citep{Ferguson:1973}.

Note that if we truncate the stick-breaking process after $L-1$
breaks, we obtain a Dirichlet distribution on an $L$-dimensional
vector.  The first $L-1$ components of this vector manifest the same
kind of bias towards larger values for earlier components as the full
stick-breaking distribution.  However, the last component $\theta_L$
represents the portion of the stick that remains after $L-1$ breaks
and has less of a bias toward small values than in the untruncated
case.

Finally, we will find it convenient to define a two-parameter variant
of the GEM distribution that allows control over both the mean and
variance of stick lengths.  We denote this distribution as $\gem(m,
\pi)$, in which $\pi > 0$ and $m \in (0,1)$.  In this variant, the
stick lengths are defined as $V_i \sim \bet(m\pi, (1-m)\pi)$.  The
standard $\gem(\gamma)$ is the special case when $m \pi = 1$ and
$\gamma = (1-m) \pi$.  Note that its mean and variance are tied
through its single parameter.

\section{The nested Chinese restaurant process}
\label{sec:nCRP}

\begin{figure}
  \begin{center}
    \includegraphics[width=0.9 \textwidth]{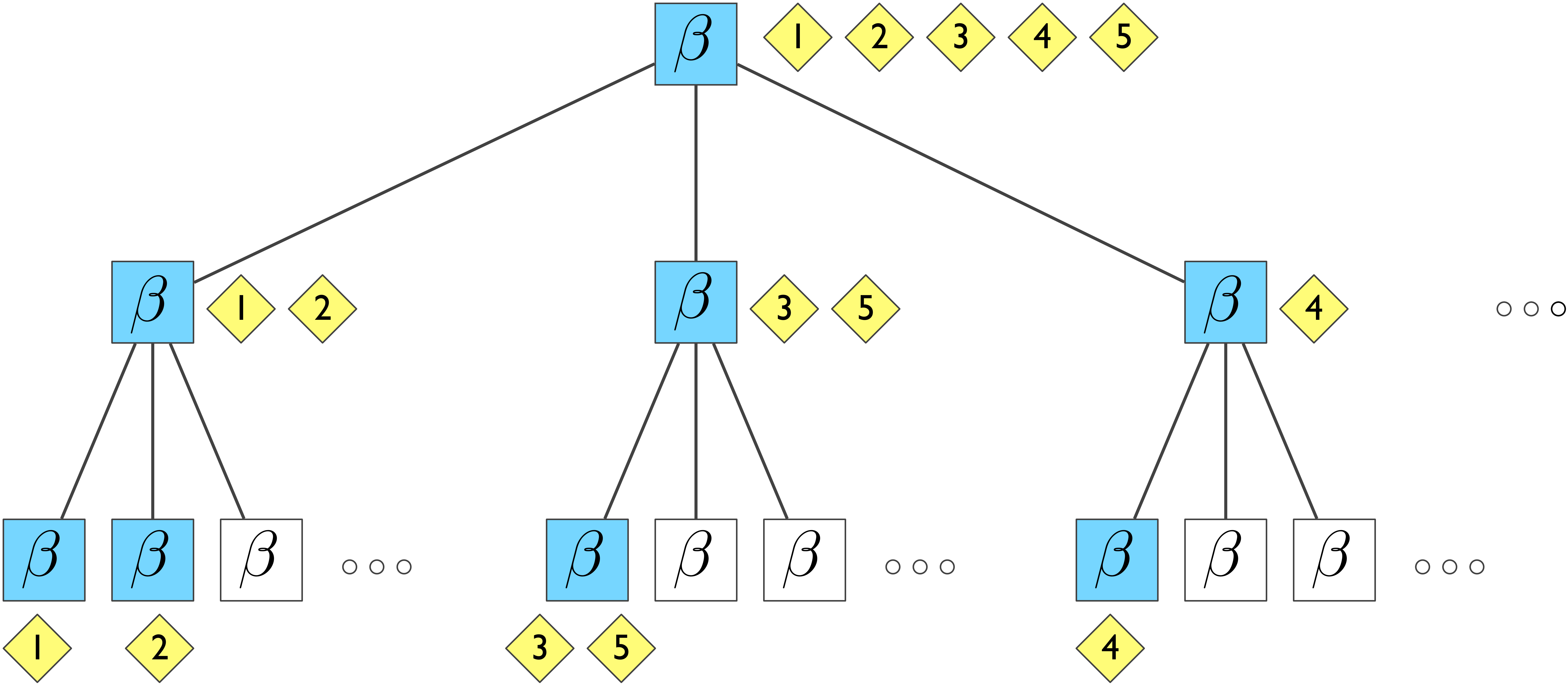}
  \end{center}
  \caption{\label{fig:nCRP} A configuration of the nested Chinese
    restaurant process illustrated to three levels.  Each box
    represents a restaurant with an infinite number of tables, each of
    which refers to a unique table in the next level of the tree.  In
    this configuration, five tourists have visited restaurants along
    four unique paths.  Their paths trace a subtree in the infinite
    tree.  (Note that the configuration of customers within each
    restaurant can be determined by observing the restaurants chosen
    by customers at the next level of the tree.)  In the hLDA model of
    \mysec{hlda}, each restaurant is associated with a topic
    distribution $\beta$.  Each document is assumed to choose its
    words from the topic distributions along a randomly chosen path.}
\end{figure}

The Chinese restaurant process and related distributions are widely
used in Bayesian nonparametric statistics because they make it
possible to define statistical models in which observations are
assumed to be drawn from an unknown number of classes. However, this
kind of model is limited in the structures that it allows to be
expressed in data.  Analyzing the richly structured data that are
common in computer science requires extending this approach. In this
section we discuss how similar ideas can be used to define a
probability distribution on infinitely-deep, infinitely-branching
trees.  This distribution is subsequently used as a prior distribution
in a hierarchical topic model that identifies documents with paths
down the tree.

A tree can be viewed as a nested sequence of partitions.  We obtain a
distribution on trees by generalizing the CRP to such sequences.
Specifically, we define a \emph{nested Chinese restaurant process}
(nCRP) by imagining the following scenario for generating a sample.
Suppose there are an infinite number of infinite-table Chinese
restaurants in a city.  One restaurant is identified as the root
restaurant, and on each of its infinite tables is a card with the name
of another restaurant.  On each of the tables in those restaurants are
cards that refer to other restaurants, and this structure repeats
infinitely many times.\footnote{A finite-depth precursor of
this model was presented in \citeN{Blei:2003}.}
Each restaurant is referred to exactly once; thus, the restaurants in
the city are organized into an infinitely-branched, infinitely-deep
tree.  Note that each restaurant is associated with a level in this
tree.  The root restaurant is at level 1, the restaurants referred to
on its tables' cards are at level 2, and so on.

A tourist arrives at the city for an culinary vacation.  On the first
evening, he enters the root Chinese restaurant and selects a table
using the CRP distribution in \myeq{CRP}.  On the second evening, he
goes to the restaurant identified on the first night's table and
chooses a second table using a CRP distribution based on the occupancy
pattern of the tables in the second night's restaurant.  He repeats
this process forever. After $M$ tourists have been on vacation in the
city, the collection of paths describes a random subtree of the
infinite tree; this subtree has a branching factor of at most $M$ at
all nodes.  See \myfig{nCRP} for an example of the first three levels
from such a random tree.

There are many ways to place prior distributions on trees, and our
specific choice is based on several considerations.  First and
foremost, a prior distribution combines with a likelihood to yield a
posterior distribution, and we must be able to compute this posterior
distribution.  In our case, the likelihood will arise from the
hierarchical topic model to be described in \mysec{hlda}.  As we will
show in \mysec{posterior}, the specific prior that we propose in this
section combines with the likelihood to yield a posterior distribution
that is amenable to probabilistic inference.  Second, we have retained
important aspects of the CRP, in particular the ``preferential
attachment'' dynamics that are built into \myeq{CRP}.  Probability
structures of this form have been used as models in a variety of
applications~\citep{Barabasi:1999,Krapivsky:2000,Albert:2001,Drinea:2006},
and the clustering that they induce makes them a reasonable starting
place for a hierarchical topic model.

In fact, these two points are intimately related.  The CRP yields an
\textit{exchangeable} distribution across partitions, i.e., the
distribution is invariant to the order of the arrival of
customers~\citep{Pitman:2002}.  This exchangeability property makes
CRP-based models amenable to posterior inference using Monte Carlo
methods~\citep{Escobar:1995,MacEachern:1998,Neal:2000}.

\section{Hierarchical latent Dirichlet allocation}
\label{sec:hlda}

The nested CRP provides a way to define a prior on tree topologies
that does not limit the branching factor or depth of the trees. We can
use this distribution as a component of a probabilistic topic
model.

The goal of topic modeling is to identify subsets of words that tend
to co-occur within documents.  Some of the early work on topic
modeling derived from latent semantic analysis, an application of the
singular value decomposition in which ``topics'' are viewed post hoc
as the basis of a low-dimensional subspace~\citep{Deerwester:1990}.
Subsequent work treated topics as probability distributions over words
and used likelihood-based methods to estimate these distributions from
a corpus~\citep{Hofmann:1999a}.  In both of these approaches, the
interpretation of ``topic'' differs in key ways from the clustering
metaphor because the same word can be given high probability (or
weight) under multiple topics.  This gives topic models the capability
to capture notions of polysemy (e.g., ``bank'' can occur with high
probability in both a finance topic and a waterways topic).
Probabilistic topic models were given a fully Bayesian treatment in
the latent Dirichlet allocation (LDA) model~\citep{Blei:2003b}.

Topic models such as LDA treat topics as a ``flat'' set of probability
distributions, with no direct relationship between one topic and
another.  While these models can be used to recover a set of topics
from a corpus, they fail to indicate the level of abstraction of a
topic, or how the various topics are related. The model that we
present in this section builds on the nCRP to define a
\emph{hierarchical topic model}.  This model arranges the topics into
a tree, with the desideratum that more general topics should appear
near the root and more specialized topics should appear near the
leaves~\citep{Hofmann:1999}.  Having defined such a model, we use
probabilistic inference to simultaneously identify the topics and the
relationships between them.

Our approach to defining a hierarchical topic model is based on
identifying documents with the paths generated by the nCRP.  We
augment the nCRP in two ways to obtain a generative model for
documents.  First, we associate a topic, i.e., a probability
distribution across words, with each node in the tree.  A path in the
tree thus picks out an infinite collection of topics.  Second, given a
choice of path, we use the GEM distribution to define a probability
distribution on the topics along this path.  Given a draw from a GEM
distribution, a document is generated by repeatedly selecting topics
according to the probabilities defined by that draw, and then drawing
each word from the probability distribution defined by its selected
topic.

More formally, consider the infinite tree defined by the nCRP and let
$\vct{c}_d$ denote the path through that tree for the $d$th customer
(i.e., document).  In the \emph{hierarchical LDA} (hLDA) model, the
documents in a corpus are assumed drawn from the following generative
process:
\begin{enumerate}
\item For each table $k \in {\cal T}$ in the infinite tree,
  \begin{enumerate}
  \item Draw a topic $\beta_{k} \sim \Dir(\eta)$.
  \end{enumerate}
\item For each document, $d \in \{1, 2, \ldots, D \}$
  \begin{enumerate}
  \item Draw $\vct{c}_d \sim \textrm{nCRP}(\gamma)$.
  \item Draw a distribution over levels in the tree, $\theta_d \g
    \{m, \pi\} \sim \gem(m, \pi)$.
  \item For each word,
    \begin{enumerate}
    \item Choose level $Z_{d,n} \g \theta \sim \mult(\theta_d)$.
    \item Choose word $W_{d,n} \g \{ z_{d,n}, \vct{c}_d,
      \vctg{\beta}\} \sim \mult(\beta_{\vct{c}_d}[z_{d,n}])$, which is
      parameterized by the topic in position $z_{d,n}$ on the path
      $\vct{c}_d$.
    \end{enumerate}
  \end{enumerate}
\end{enumerate}
This generative process defines a probability distribution across
possible corpora.

The goal of finding a topic hierarchy at different levels of
abstraction is distinct from the problem of hierarchical
clustering~\cite{Zamir:1998,Larsen:1999,Vaithyanathan:2000,Duda:2000,Hastie:2001,Heller:2005}. Hierarchical
clustering treats each data point as a leaf in a tree, and merges
similar data points up the tree until all are merged into a root node.
Thus, internal nodes represent summaries of the data below which, in
this setting, would yield distributions across words that share high
probability words with their children.

In the hierarchical topic model, the internal nodes are not summaries
of their children.  Rather, the internal nodes reflect the shared
terminology of the documents assigned to the paths that contain them.
This can be seen in \myfig{jacm}, where the high probability words of
a node are distinct from the high probability words of its children.

It is important to emphasize that our approach is an unsupervised
learning approach in which the probabilistic components that we have
defined are latent variables.  That is, we do not assume that topics
are predefined, nor do we assume that the nested partitioning of
documents or the allocation of topics to levels are predefined.  We
infer these entities from a Bayesian computation in which a posterior 
distribution is obtained from conditioning on a corpus and computing 
probabilities for all latent variables.

As we will see experimentally, there is statistical pressure in the
posterior to place more general topics near the root of the tree and
to place more specialized topics further down in the tree.  To see
this, note that each path in the tree includes the root node.  Given
that the GEM distribution tends to assign relatively large
probabilities to small integers, there will be a relatively large
probability for documents to select the root node when generating
words. Therefore, to explain an observed corpus, the topic at the root
node will place high probability on words that are useful across all
the documents.

Moving down in the tree, recall that each document is assigned to a
single path.  Thus, the first level below the root induces a coarse
partition on the documents, and the topics at that level will place
high probability on words that are useful within the corresponding
subsets.  As we move still further down, the nested partitions of
documents become finer.  Consequently, the corresponding topics will
be more specialized to the particular documents in those paths.

We have presented the model as a two-phase process: an infinite set of
topics are generated and assigned to all of the nodes of an infinite
tree, and then documents are obtained by selecting nodes in the tree
and drawing words from the corresponding topics.  It is also possible,
however, to conceptualize a ``lazy'' procedure in which a topic is
generated only when a node is first selected.  In particular, consider
an empty tree (i.e., containing no topics) and consider generating the
first document.  We select a path and then repeatedly select nodes
along that path in order to generate words.  A topic is generated at a
node when that node is first selected and subsequent selections of the
node reuse the same topic.

After $n$ words have been generated, at most $n$ nodes will have been
visited and at most $n$ topics will have been generated.  The
$(n+1)$th word in the document can come from one of
previously generated topics or it can come from a new topic.
Similarly, suppose that $d$ documents have previously been generated.
The $(d+1)$th document can follow one of the paths laid down by an
earlier document and select only ``old'' topics, or it can branch off
at any point in the tree and generated ``new'' topics along the new
branch.

This discussion highlights the nonparametric nature of our model.
Rather than describing a corpus by using a probabilistic model
involving a fixed set of parameters, our model assumes that the number
of parameters can grow as the corpus grows, both within documents and
across documents.  New documents can spark new subtopics or new
specializations of existing subtopics.  Given a corpus, this
flexibility allows us to use approximate posterior inference to
discover the particular tree of topics that best describes its
documents.

It is important to note that even with this flexibility, the model
still makes assumptions about the tree. Its size, shape, and character
will be affected by the settings of the hyperparameters.  The most
influential hyperparameters in this regard are the Dirichlet parameter
for the topics $\eta$ and the stick-breaking parameters for the topic
proportions $\{m,\pi\}$.  The Dirichlet parameter controls the
sparsity of the topics; smaller values of $\eta$ will lead to topics
with most of their probability mass on a small set of words.  With a
prior bias to sparser topics, the posterior will prefer more topics to
describe a collection and thus place higher probability on larger
trees.  The stick-breaking parameters control how many words in the
documents are likely to come from topics of varying abstractions.  If
we set $\pi$ to be large (e.g., $\pi=0.5$) then the posterior will
more likely assign more words from each document to higher levels of
abstraction.  Setting $m$ to be large (e.g., $m=100$) means that word
allocations will not likely deviate from such a setting.

How we set these hyperparameters depends on the goal of the analysis.
When we analyze a document collection with hLDA for discovering and
visualizing a hierarchy embedded within it, we might examine various
settings of the hyperparameters to find a tree that meets our
exploratory needs.  We analyze documents with this purpose in mind
in \mysec{abstracts}.  In a different setting, when we are looking for a
good predictive model of the data, e.g., to compare hLDA to other
statistical models of text, then it makes sense to ``fit'' the
hyperparameters by placing priors on them and computing their
posterior.  We describe posterior inference for the hyperparameters in
\mysec{hyperparameter} and analyze documents using this approach in
\mysec{quant}.

Finally, we note that hLDA is the simplest model that exploits the
nested CRP, i.e., a flexible hierarchy of distributions, in the topic
modeling framework.  In a more complicated model, one could consider a
variant of hLDA where each document exhibits multiple paths through
the tree.  This can be modeled using a two-level distribution for word
generation: first choose a path through the tree, and then choose a
level for the word.

Recent extensions to topic models can also be adapted to make use of a
flexible topic hierarchy.  As examples, in the dynamic topic model the
documents are time stamped and the underlying topics change over
time~\citep{Blei:2006a}; in the author-topic model the authorship of
the documents affects which topics they
exhibit~\citep{Rosen-Zvi:2004}.  This said, some extensions are more
easily adaptable than others.  In the correlated topic model, the
topic proportions exhibit a covariance structure~\citep{Blei:2007}.
This is achieved by replacing a Dirichlet distribution with a logistic
normal, and the application of Bayesian nonparametric extensions is
less direct.

\subsection{Related work}

In previous work, researchers have developed a number of methods that
employ hierarchies in analyzing text data.  In one line of work, the
algorithms are given a hierarchy of document categories, and their
goal is to correctly place documents within
it~\citep{Koller:1997,Chakrabarti:1998,McCallum:1999,Dumais:2000}.
Other work has focused on deriving hierarchies of individual terms
using side information, such as a grammar or a thesaurus, that are
sometimes available for text
domains~\citep{Sanderson:1999,Stoica:2004,Cimiano:2005}.

Our method provides still another way to employ a notion of hierarchy
in text analysis.  First, rather than learn a hierarchy of terms we
learn a hierarchy of topics, where a topic is a distribution over
terms that describes a significant pattern of word co-occurrence in
the data.  Moreover, while we focus on text, a ``topic'' is simply a
data-generating distribution; we do not rely on any text-specific side
information such as a thesaurus or grammar.  Thus, by using other data
types and distributions, our methodology is readily applied to
biological data sets, purchasing data, collections of images, or
social network data.  (Note that applications in such domains have
already been demonstrated for flat topic
models~\citep{Pritchard:2000,Marlin:2003,Fei-Fei:2005,Blei:2003a,Airoldi:2008}.)
Finally, as a Bayesian nonparametric model, our approach can
accommodate future data that might lie in new and previously
undiscovered parts of the tree.  Previous work commits to a single
fixed tree for all future data.

\section{Probabilistic inference}
\label{sec:posterior}

With the hLDA model in hand, our goal is to perform \textit{posterior
inference}, i.e., to ``invert'' the generative process of documents
described above for estimating the hidden topical structure of a
document collection.  We have constructed a joint distribution of
hidden variables and observations---the latent topic structure and
observed documents---by combining prior expectations about the kinds
of tree topologies we will encounter with a generative process for
producing documents given a particular topology. We are now interested
in the distribution of the hidden structure \textit{conditioned} on
having seen the data, i.e., the distribution of the underlying topic
structure that might have generated an observed collection of
documents.  Finding this posterior distribution for different kinds of
data and models is a central problem in Bayesian statistics.  See
~\citeN{Bernardo:1994} and \citeN{Gelman:1995} for general
introductions to Bayesian statistics.

In our nonparametric setting, we must find a posterior distribution on
countably infinite collections of objects---hierarchies, path
assignments, and level allocations of words---given a collection of
documents.  Moreover, we need to be able to do this using the finite
resources of the computer.  Not surprisingly, the posterior
distribution for hLDA is not available in closed form.  We must appeal
to an approximation.

We develop a Markov chain Monte Carlo (MCMC) algorithm to approximate
the posterior for hLDA. In MCMC, one samples from a target
distribution on a set of variables by constructing a Markov chain that
has the target distribution as its stationary
distribution~\citep{Robert:2004}.  One then samples from the chain for
sufficiently long that it approaches the target, collects the sampled
states thereafter, and uses those collected states to estimate the
target.  This approach is particularly straightforward to apply to
latent variable models, where we take the state space of the Markov
chain to be the set of values that the latent variables can take on,
and the target distribution is the conditional distribution of these
latent variables given the observed data.

The particular MCMC algorithm that we present in this paper is a
\textit{Gibbs sampling} algorithm~\citep{Geman:1984,Gelfand:1990}.  In
a Gibbs sampler each latent variable is iteratively sampled
conditioned on the observations and all the other latent variables.
We employ \emph{collapsed Gibbs sampling}~\citep{Liu:1994}, in which
we marginalize out some of the latent variables to speed up the
convergence of the chain.  Collapsed Gibbs sampling for topic
models~\citep{Griffiths:2004a} has been widely used in a number of
topic modeling applications~\citep{McCallum:2004,Rosen-Zvi:2004,Mimno:2007,Dietz:2007,Newman:2007}.

\begin{figure}
  \begin{center}
    \includegraphics[width=0.9\textwidth]{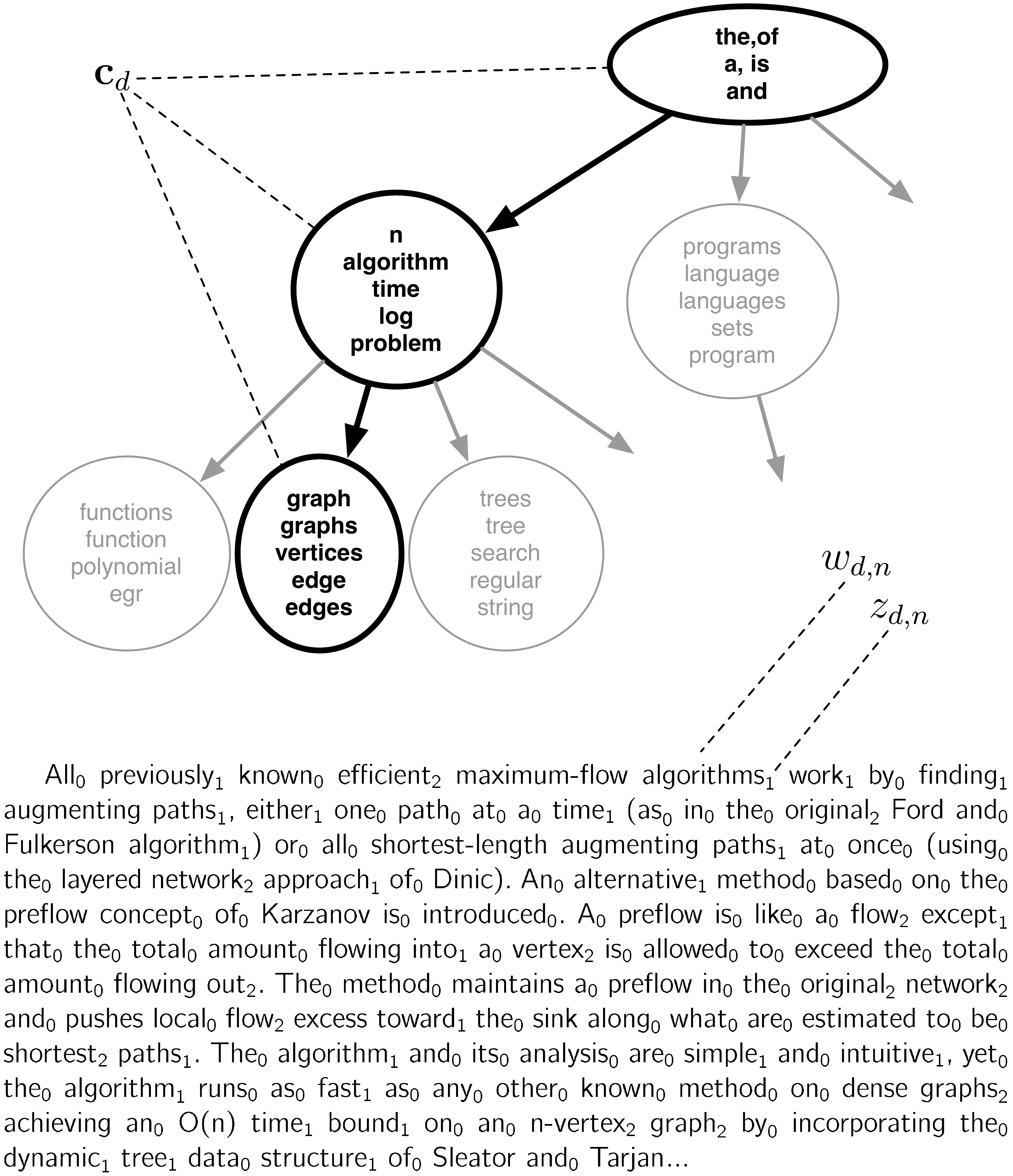}
  \end{center}
  \caption{\label{fig:gibbs-state} A single state of the Markov chain in
    the Gibbs sampler for the abstract of ``A new approach to the
    maximum-flow problem'' [Goldberg and Tarjan, 1986].  The document is
    associated with a path through the hierarchy $\textbf{c}_d$, and each
    node in the hierarchy is associated with a distribution over terms.
    (The five most probable terms are illustrated.)  Finally, each word in
    the abstract $w_{d,n}$ is associated with a level in the path through
    the hierarchy $z_{d,n}$, with $0$ being the highest level and $2$
    being the lowest.  The Gibbs sampler iteratively draws $\textbf{c}_d$
    and $z_{d,n}$ for all words in all documents (see \mysec{posterior}).}

%    The assignments illustrated here were taken at the approximate mode of
%    the posterior when analyzing the JACM corpus (see Section XXX)

\end{figure}

In hLDA, we sample the per-document paths $\vct{c}_d$
and the per-word level allocations to topics in those paths $z_{d,n}$.
We marginalize out the topic parameters $\beta_i$ and the 
per-document topic proportions $\theta_d$.  The state of the
Markov chain is illustrated, for a single document, in
\myfig{gibbs-state}.  (The particular assignments illustrated in the
figure are taken at the approximate mode of the hLDA model posterior
conditioned on abstracts from the JACM.)

Thus, we approximate the posterior $p(\vct{c}_{1:D}, \vct{z}_{1:D} \g
\gamma, \eta, m, \pi, \vct{w}_{1:D})$.  The hyperparameter $\gamma$
reflects the tendency of the customers in each restaurant to share
tables, $\eta$ reflects the expected variance of the underlying topics
(e.g, $\eta \ll 1$ will tend to choose topics with fewer
high-probability words), and $m$ and $\pi$ reflect our expectation
about the allocation of words to levels within a document.  The
hyperparameters can be fixed according to the constraints of the
analysis and prior expectation about the data, or inferred as
described in \mysec{hyperparameter}.

Intuitively, the CRP parameter $\gamma$ and topic prior $\eta$ provide
control over the size of the inferred tree.  For example, a model with
large $\gamma$ and small $\eta$ will tend to find a tree with more
topics.  The small $\eta$ encourages fewer words to have high
probability in each topic; thus, the posterior requires more topics to
explain the data. The large $\gamma$ increases the likelihood that
documents will choose new paths when traversing the nested CRP.

The GEM parameter $m$ reflects the proportion of general words
relative to specific words, and the GEM parameter $\pi$ reflects how
strictly we expect the documents to adhere to these proportions.  A
larger value of $\pi$ enforces the notions of generality and
specificity that lead to more interpretable trees.

The remainder of this section is organized as follows.  First, we
outline the two main steps in the algorithm: the sampling of level
allocations and the sampling of path assignments.  We then combine
these steps into an overall algorithm.  Next, we present prior
distributions for the hyperparameters of the model and describe
posterior inference for the hyperparameters.  Finally, we outline how
to assess the convergence of the sampler and approximate the mode of
the posterior distribution.

\subsection{Sampling level allocations}

Given the current path assignments, we need to sample the level
allocation variable $z_{d,n}$ for word $n$ in document $d$ from its
distribution given the current values of all other variables:
\begin{equation}
  \label{eq:topicassignment}
  p(z_{d,n} \g \vct{z}_{-(d,n)}, \vct{c},
  \vct{w}, m, \pi, \eta) \propto
  p(z_{d,n} \g \vct{z}_{d,-n}, m, \pi)
  p(w_{d,n} \g \vct{z}, \vct{c}, \vct{w}_{-(d,n)}, \eta),
\end{equation}
where $\vct{z}_{-(d,n)}$ and $\vct{w}_{-(d,n)}$ are the vectors of
level allocations and observed words leaving out $z_{d,n}$ and
$w_{d,n}$ respectively. We will use similar notation whenever items
are left out from an index set; for example, $\vct{z}_{d,-n}$ denotes
the level allocations in document $d$, leaving out $z_{d,n}$.

The first term in \myeq{topicassignment} is a distribution over
levels.  This distribution has an infinite number of components, so we
sample in stages.  First, we sample from the distribution over the
space of levels that are currently represented in the rest of the
document, i.e., $\max(\vct{z}_{d,-n})$, and a level deeper than that
level.  The first components of this distribution are, for $k \leq
\max(\vct{z}_{d,-n})$,
\begin{eqnarray*}
  p(z_{d,n} = k \g \vct{z}_{d, -n}, m, \pi) &=&
  \E\left[V_k \prod_{j=1}^{k-1} V_{j} \g \vct{z}_{d, -n}, m, \pi\right] \\
  &=& \E[V_k \g \vct{z}_{d,-n}, m, \pi] \prod_{j=1}^{k-1} \E[1 - V_{j} \g
  \vct{z}_{d,-n}, m, \pi] \\
  &=& \frac{(1-m) \pi + \cnt{\vct{z}_{d,-n} = k}}{\pi + \cnt{\vct{z}_{d,-n} \geq k}}
  \prod_{j=1}^{k-1} \frac{m\pi + \cnt{\vct{z}_{d,-n} >j}}{\pi +
    \cnt{\vct{z}_{d,-n}\geq j}}
\end{eqnarray*}
where $\cnt{\cdot}$ counts the elements of an array satisfying a given
condition.

The second term in \myeq{topicassignment} is the probability of a given
word based on a possible assignment. From the assumption that
the topic parameters $\beta_i$ are generated from a Dirichlet distribution
with hyperparameters $\eta$ we obtain
\begin{equation}
p(w_{d,n} \g \vct{z}, \vct{c}, \vct{w}_{-(d,n)}, \eta) \propto
\cnt{\vct{z}_{-(d,n)}=z_{d,n},\vct{c}_{z_{d,n}}=
  c_{d,z_{d,n}}, \vct{w}_{-(d,n)}=w_{d,n}}+\eta
% {\cnt{
% \vct{z}_{-(d,n)}=z_{d,n},\vct{c}={\bf c}_d}+V\eta},
% above: cut denominator
\end{equation}
which is the smoothed frequency of seeing word $w_{d,n}$ allocated to
the topic at level $z_{d,n}$ of the path ${\bf c}_d$.

The last component of the distribution over topic assignments is
\begin{equation*}
  p(z_{d,n} > \max(\vct{z}_{d,-n}) \g \vct{z}_{d,-n}, \vct{w}, m, \pi,
  \eta) =
  1 - \sum_{j=1}^{\max(\vct{z}_{d,-n})} p(z_{d,n} = j \g
  \vct{z}_{d,-n}, \vct{w}, m, \pi, \eta).
\end{equation*}
If the last component is sampled then we sample from a Bernoulli
distribution for increasing values of $\ell$, starting with $\ell =
\max(\vct{z}_{d,-n}) + 1$, until we determine $z_{d,n}$,
\begin{eqnarray*}
  p(z_{d,n} = \ell \g z_{d,-n}, z_{d,n} > \ell-1, \vct{w}, m, \pi, \eta)
  &=& (1-m) p(w_{d,n} \g \vct{z}, \vct{c}, \vct{w}_{-(d,n)}, \eta) \\
  p(z_{d,n} > \ell \g z_{d,-n}, z_{d,n} > \ell-1) &=& 1 - p(z_{d,n} =
  \ell \g z_{d,-n}, z_{d,n} > \ell-1, \vct{w}, m, \pi, \eta).
\end{eqnarray*}
Note that this changes the maximum level when resampling subsequent
level assignments.

% Let $n_{\vct{z}}$ denote the $L$-vector of topic counts in $\vct{z}$,
% and let $n_{(\vct{w}, \vct{z}, \vct{c})}$ denote the word counts
% assigned to the tree defined by the path assignments.  The terms in
% \myeq{topicassignment} are
% \begin{eqnarray*}
%   p(z_{m,n} \g \vct{z}_{m,-n}) &\propto& \alpha + n_{\vct{z}_{-m}}^{z_{m,n}} \\
%   p(w_{m,n} \g \vct{z}, \vct{c}, \vct{w}_{-(m,n)}) &\propto&
%   \eta + n_{(\vct{w}_{-(m,n)}, \vct{z}_{-(m,n)}, \vct{c})}^{c_{z_{m,n}}, w_{m,n}}.
% \end{eqnarray*}

\subsection{Sampling paths}

Given the level allocation variables, we need to sample the path
associated with each document conditioned on all other paths and the
observed words.  We appeal to the fact that $\max(\vct{z}_{d})$ is finite,
and are only concerned with paths of that length:
\begin{equation}
  \label{eq:topicpath}
  p(\vct{c}_d \g \vct{w}, \vct{c}_{-d}, \vct{z}, \eta, \gamma) \propto
  p(\vct{c}_d \g \vct{c}_{-d}, \gamma)
  p(\vct{w}_d \g \vct{c}, \vct{w}_{-d}, \vct{z}, \eta).
\end{equation}
This expression is an instance of Bayes's theorem with $p(\vct{w}_d \g
\cc, \vct{w}_{-d}, \vct{z}, \eta)$ as the probability of the data given
a particular choice of path, and $p(\cc_d \g \cc_{-d}, \gamma)$ as the
prior on paths implied by the nested CRP. The probability of the data 
is obtained by integrating over the multinomial parameters, which gives 
a ratio of normalizing constants for the Dirichlet distribution,
\begin{small}
\begin{equation*}
  \begin{split}
    & p(\vct{w}_d \g \cc, \vct{w}_{-d}, \vct{z}, \eta) = \\
    & \prod_{\ell = 1}^{\max(\vct{z}_d)}
    \frac
    {\Gamma\left(
        \textstyle \sum_w
        \cnt{\vct{z}_{-d}=\ell, \vct{c}_{-d,\ell}=c_{d,\ell}, \vct{w}_{-d}=w}
        + V \eta
      \right)}
    {\prod_w \Gamma\left(
        \cnt{\vct{z}_{-d}=\ell, \vct{c}_{-d, \ell}=c_{d,\ell}, \vct{w}_{-d}=w}
        + \eta
      \right)}
    \frac
    {\prod_w \Gamma\left(
        \cnt{\vct{z}=\ell, \vct{c}_{\ell}=c_{d, \ell}, \vct{w}=w}
        + \eta
      \right)}
    {\Gamma\left(\textstyle \sum_w
        \cnt{\vct{z}=\ell, \vct{c}_{\ell}=c_{d, \ell}, \vct{w}=w}
        + V \eta
      \right)},
  \end{split}
\end{equation*}
\end{small}

\noindent where we use the same notation for counting over arrays of
variables as above. Note that the path must be drawn as a block,
because its value at each level depends on its value at the previous
level.  The set of possible paths corresponds to the union of the set
of existing paths through the tree, each represented by a leaf, with
the set of possible novel paths, each represented by an internal node.

\subsection{Summary of Gibbs sampling algorithm}

With these conditional distributions in hand, we specify the full
Gibbs sampling algorithm.  Given the current state of the sampler,
$\{\vct{c}^{(t)}_{1:D}, \vct{z}^{(t)}_{1:D}\}$, we iteratively sample
each variable conditioned on the rest.
\begin{packed_enumerate}
\item For each document $d \in \{1, \ldots, D\}$
  \begin{packed_enumerate}
  \item Randomly draw $\vct{c}_d^{(t+1)}$ from \myeq{topicpath}.
  \item Randomly draw $z_{n,d}^{(t+1)}$ from \myeq{topicassignment}
      for each word, $n \in \{1, \ldots N_d\}$.
  \end{packed_enumerate}
\end{packed_enumerate}
The stationary distribution of the corresponding Markov chain is the
conditional distribution of the latent variables in the hLDA model
given the corpus.  After running the chain for sufficiently many
iterations that it can approach its stationary distribution (the
``burn-in'') we can collect samples at intervals selected to minimize
autocorrelation, and approximate the true posterior with the
corresponding empirical distribution.

Although this algorithm is guaranteed to converge in the limit, it is
difficult to say something more definitive about the speed of the
algorithm independent of the data being analyzed.  In hLDA, we sample
a leaf from the tree for each document $\textbf{c}_d$ and a level
assignment for each word $z_{d,n}$.  As described above, the number of
items from which each is sampled depends on the current state of the
hierarchy and other level assignments in the document.  Two data sets
of equal size may induce different trees and yield different running
times for each iteration of the sampler.  For the corpora analyzed
below in \mysec{abstracts}, the Gibbs sampler averaged 0.001 seconds
per document for the JACM data and \emph{Psychological Review} data,
and 0.006 seconds per document for the \emph{Proceedings of the
  National Academy of Sciences} data.\footnote{Timings were measured
  with the Gibbs sampler running on a 2.2GHz Opteron 275 processor.}

\subsection{Sampling the hyperparameters}
\label{sec:hyperparameter}

The values of hyperparameters are generally unknown a priori.  We
include them in the inference process by endowing them with prior
distributions,
\begin{eqnarray*}
  m &\sim& \textrm{Beta}(\alpha_1, \alpha_2) \\
  \pi &\sim& \textrm{Exponential}(\alpha_3) \\
  \gamma &\sim& \textrm{Gamma}(\alpha_4, \alpha_5) \\
  \eta &\sim& \textrm{Exponential}(\alpha_6).
\end{eqnarray*}
These priors also contain parameters (``hyper-hyperparameters''), 
but the resulting inferences are less influenced by these 
hyper-hyperparameters than they are by fixing the original 
hyperparameters to specific values~\cite{Bernardo:1994}.

To incorporate this extension into the Gibbs sampler, we interleave
Metropolis-Hastings (MH) steps between iterations of the Gibbs sampler
to obtain new values of $m$, $\pi$, $\gamma$, and $\eta$.  This
preserves the integrity of the Markov chain, although it may mix
slower than the collapsed Gibbs sampler without the MH
updates~\citep{Robert:2004}.

\subsection{Assessing convergence and approximating the mode}
\label{sec:convergence}

\begin{figure}
  \begin{center}
    \includegraphics[width=\textwidth]{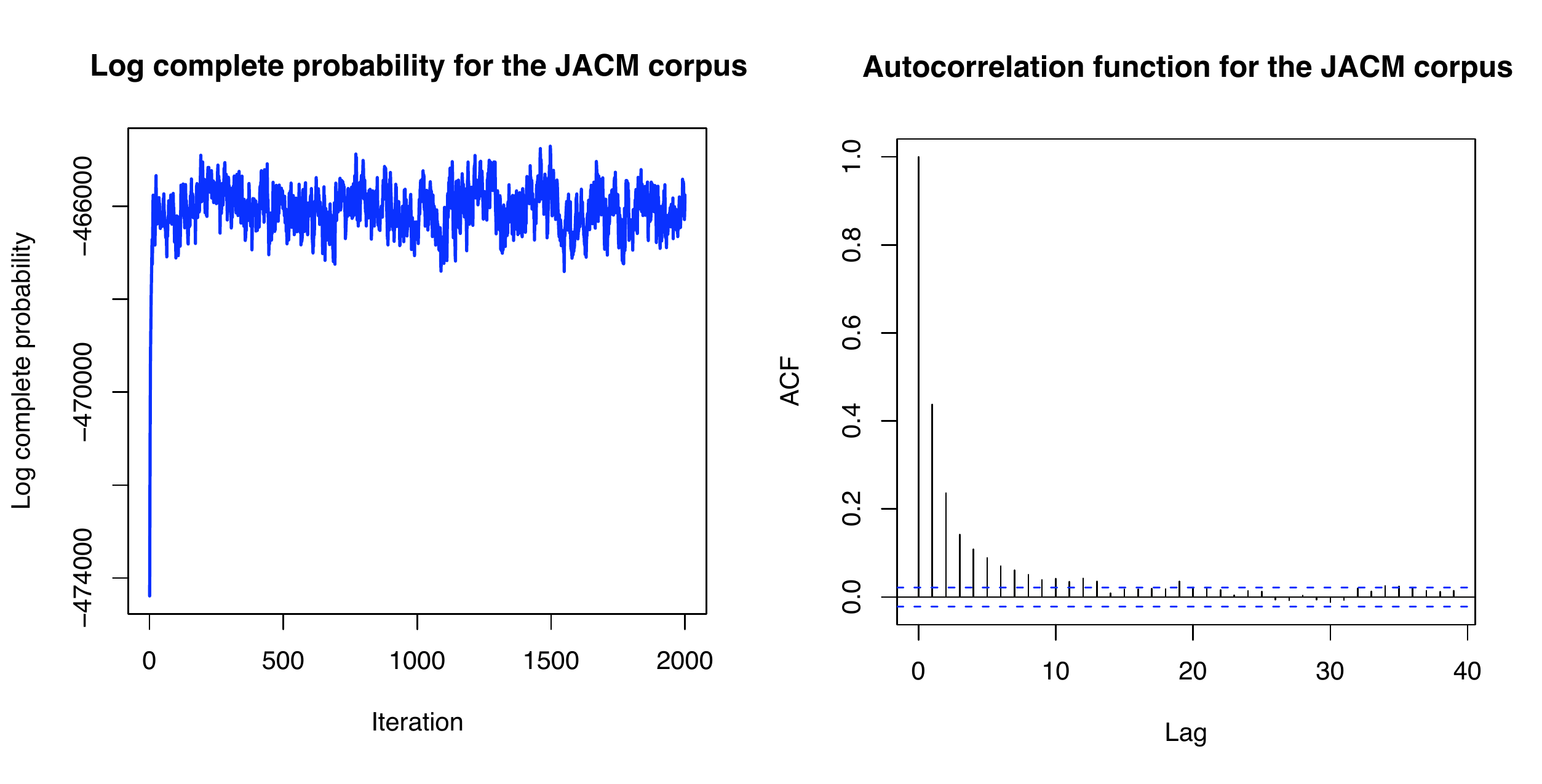}
  \end{center}
  \caption{\label{fig:ACF} (Left) The complete log likelihood of
    \myeq{score} for the first 2000 iterations of the Gibbs sampler
    run on the JACM corpus of \mysec{abstracts}.  (Right) The
    autocorrelation function (ACF) of the log complete log likelihood
    (with confidence interval) for the remaining 8000 iterations.  The
    autocorrelation decreases rapidly as a function of the lag between
    samples.}
\end{figure}

Practical applications must address the issue of approximating the
mode of the distribution on trees and assessing convergence of the
Markov chain.  We can obtain information about both by examining the
log probability of each sampled state.  For a particular sample, i.e.,
a configuration of the latent variables, we compute the log
probability of that configuration and observations, conditioned on the
hyperparameters:
\begin{equation}
  \label{eq:score}
  {\cal L}^{(t)} = \log
  p(\vct{c}^{(t)}_{1:D}, \vct{z}^{(t)}_{1:D}, \vct{w}_{1:D} \g \gamma,
  \eta, m, \pi).
\end{equation}
With this statistic, we can approximate the mode of the posterior by
choosing the state with the highest log probability.  Moreover, we can
assess convergence of the chain by examining the autocorrelation of
${\cal L}^{(t)}$.  \myfig{ACF} (right) illustrates the autocorrelation
as a function of the number of iterations between samples (the
``lag'') when modeling the JACM corpus described in \mysec{abstracts}.
The chain was run for 10,000 iterations; 2000 iterations were
discarded as burn-in.

\myfig{ACF} (left) illustrates \myeq{score} for the burn-in
iterations.  Gibbs samplers stochastically climb the posterior
distribution surface to find an area of high posterior probability,
and then explore its curvature through sampling.  In practice, one
usually restarts this procedure a handful of times and chooses the
local mode which has highest posterior likelihood~\citep{Robert:2004}.

Despite the lack of theoretical guarantees, Gibbs sampling is
appropriate for the kind of data analysis for which hLDA and many
other latent variable models are tailored.  Rather than try to
understand the full surface of the posterior, the goal of latent
variable modeling is to find a useful representation of complicated
high-dimensional data, and a local mode of the posterior found by
Gibbs sampling often provides such a representation.  In the next
section, we will assess hLDA qualitatively, through visualization of
summaries of the data, and quantitatively, by using the latent
variable representation to provide a predictive model of text.

\section{Examples and empirical results}
\label{sec:results}

We present experiments analyzing both simulated and real text data to
demonstrate the application of hLDA and its corresponding Gibbs
sampler.

\subsection{Analysis of simulated data}
\label{sec:sim}

\begin{figure}[t]
  \begin{center}
    \rotatebox{90}{\includegraphics*[height=0.80\textwidth]{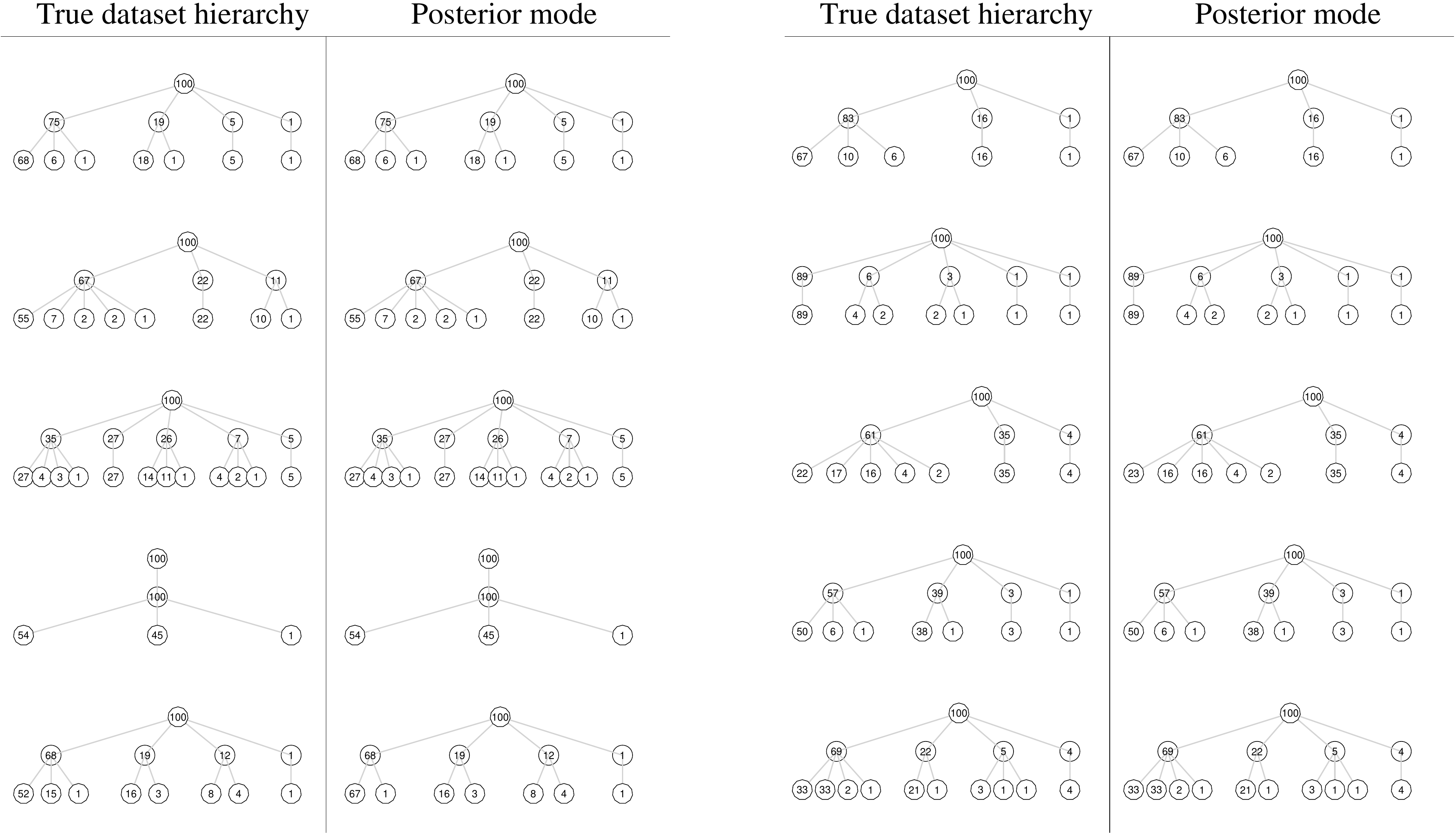}}
  \end{center}
  \caption{\label{fig:simulated-data} Inferring the mode of the
    posterior hierarchy from simulated data.  See \mysec{sim}.}
\end{figure}

In \myfig{simulated-data}, we depict the hierarchies and allocations
for ten simulated data sets drawn from an hLDA model.  For each
data set, we draw 100 documents of 250 words each.  The vocabulary size
is 100, and the hyperparameters are fixed at $\eta = .005$, and
$\gamma = 1$.  In these simulations, we truncated the stick-breaking
procedure at three levels, and simply took a Dirichlet distribution
over the proportion of words allocated to those levels. The resulting
hierarchies shown in \myfig{simulated-data} illustrate the range of
structures on which the prior assigns probability.

% TLG: what are the hyperparameters of the Dirichlet?

In the same figure, we illustrate the estimated mode of the posterior
distribution across the hierarchy and allocations for the ten
data sets.  We exactly recover the correct hierarchies, with only two
errors.  In one case, the error is a single wrongly allocated path.
In the other case, the inferred mode has higher posterior probability
than the true tree structure (due to finite data).

In general we cannot expect to always find the exact tree.  This is
dependent on the size of the data set, and how identifiable the topics
are.  Our choice of small $\eta$ yields topics that are relatively
sparse and (probably) very different from each other.  Trees will not
be as easy to identify in data sets which exhibit polysemy and
similarity between topics.

\subsection{Hierarchy discovery in scientific abstracts}
\label{sec:abstracts}

Given a document collection, one is typically interested in examining
the underlying tree of topics at the mode of the posterior.  As
described above, our inferential procedure yields a tree structure by
assembling the unique subset of paths contained in $\{\vct{c}_1,
\ldots, \vct{c}_D\}$ at the approximate mode of the posterior.

For a given tree, we can examine the topics that populate the tree.
Given the assignment of words to levels and the assignment of
documents to paths, the probability of a particular word at a
particular node is roughly proportional to the number of times that
word was generated by the topic at that node.  More specifically, the
mean probability of a word $w$ in a topic at level $\ell$ of path
${\bf p}$ is given by
\begin{equation}
  \label{eq:topic-prob}
  p(w \g \vct{z}, \vct{c}, \vct{w}, \eta) =
  \frac{\cnt{\vct{z}=\ell,\vct{c}={\bf
        p},\vct{w}=w}+\eta}{\cnt{
      \vct{z}=\ell,\vct{c}={\bf p}}+V\eta}.
\end{equation}

Using these quantities, the hLDA model can be used for
analyzing collections of scientific abstracts, recovering the
underlying hierarchical structure appropriate to a collection, and
visualizing that hierarchy of topics for a better understanding of the
structure of the corpora.  We demonstrate the analysis of three different
collections of journal abstracts under hLDA.

In these analyses, as above, we truncate the stick-breaking procedure
at three levels, facilitating visualization of the results. The topic
Dirichlet hyperparameters were fixed at $\eta = \{2.0, 1.0, 0.5\}$,
which encourages many terms in the high-level distributions, fewer
terms in the mid-level distributions, and still fewer terms in the
low-level distributions.  The nested CRP parameter $\gamma$ was fixed
at 1.0.  The GEM parameters were fixed at $m = 100$ and $\pi = 0.5$.
This strongly biases the level proportions to place more mass at the
higher levels of the hierarchy.

In \myfig{jacm}, we illustrate the approximate posterior mode of a
hierarchy estimated from a collection of 536 abstracts from the JACM.
The tree structure illustrates the ensemble of paths assigned to the
documents.  In each node, we illustrate the top five words sorted by
expected posterior probability, computed from \myeq{topic-prob}.
Several leaves are annotated with document titles.  For each leaf, we
chose the five documents assigned to its path that have the highest
numbers of words allocated to the bottom level.

The model has found the function words in the data set, assigning words
like ``the,'' ``of,'' ``or,'' and ``and'' to the root topic.  In its
second level, the posterior hierarchy appears to have captured some
of the major subfields in computer science, distinguishing between
databases, algorithms, programming languages and networking.  In
the third level, it further refines those fields.  For example, it
delineates between the verification area of networking and the
queuing area.

\begin{figure}
  \begin{center}
    \includegraphics[angle=90, width=0.7 \textheight]{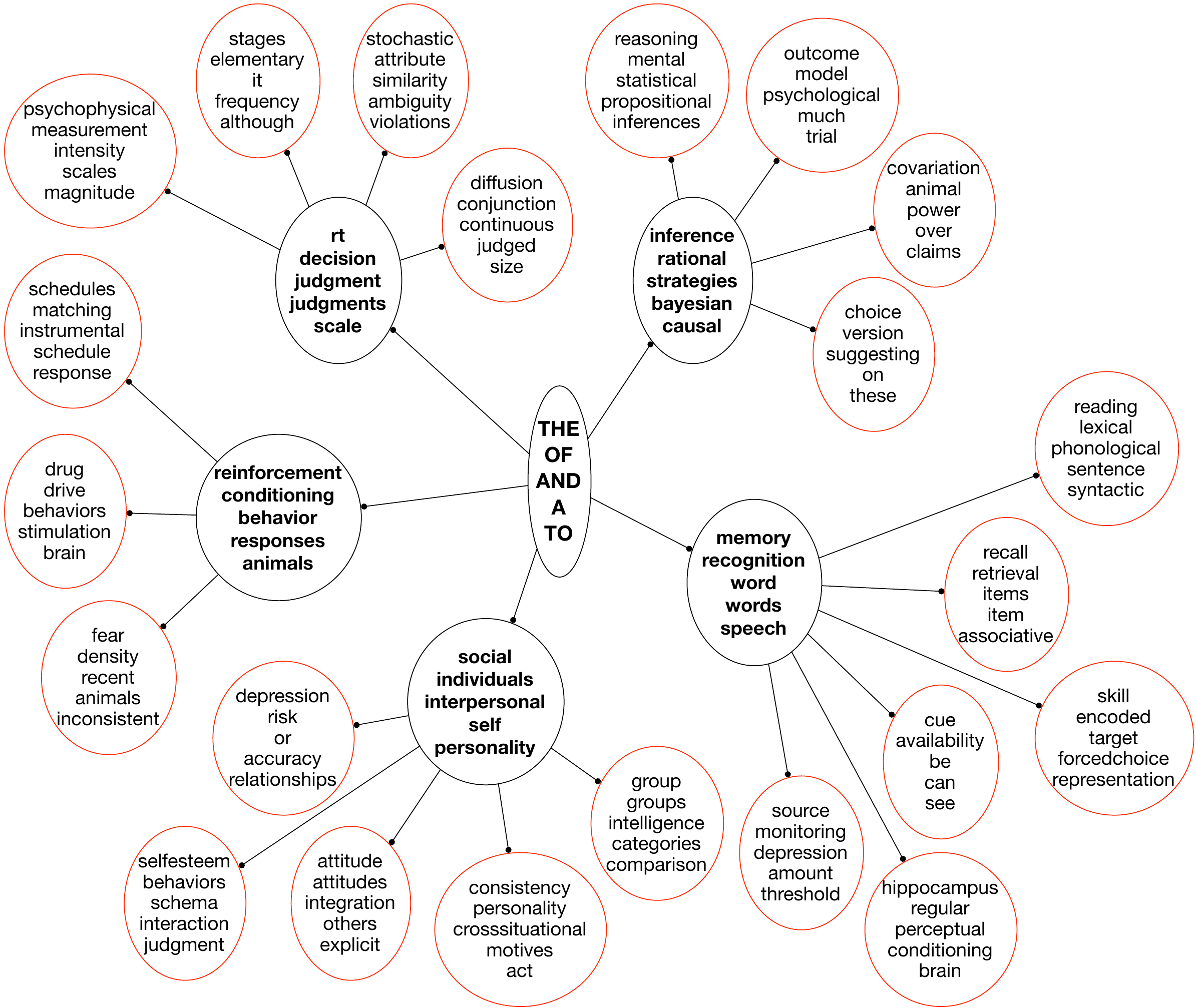}
  \end{center}
  \caption{\label{fig:psych} A portion of the hierarchy learned from
    the 1,272 abstracts of \emph{Psychological Review} from
    1967--2003.  The vocabulary was restricted to the 1,971 terms that
    occurred in more than five documents, yielding a corpus of 136K
    words.  The learned hierarchy, of which only a portion is
    illustrated, contains 52 topics.}
\end{figure}

In \myfig{psych}, we illustrate an analysis of a collection of 1,272
psychology abstracts from {\em Psychological Review} from 1967 to 2003.
Again, we have discovered an underlying hierarchical structure of the
field.  The top node contains the function words; the second level
delineates between large subfields such as behavioral, social and
cognitive psychology; the third level further refines those subfields.

\begin{figure}
  \begin{center}
    \includegraphics[angle=90, width=0.7 \textheight]{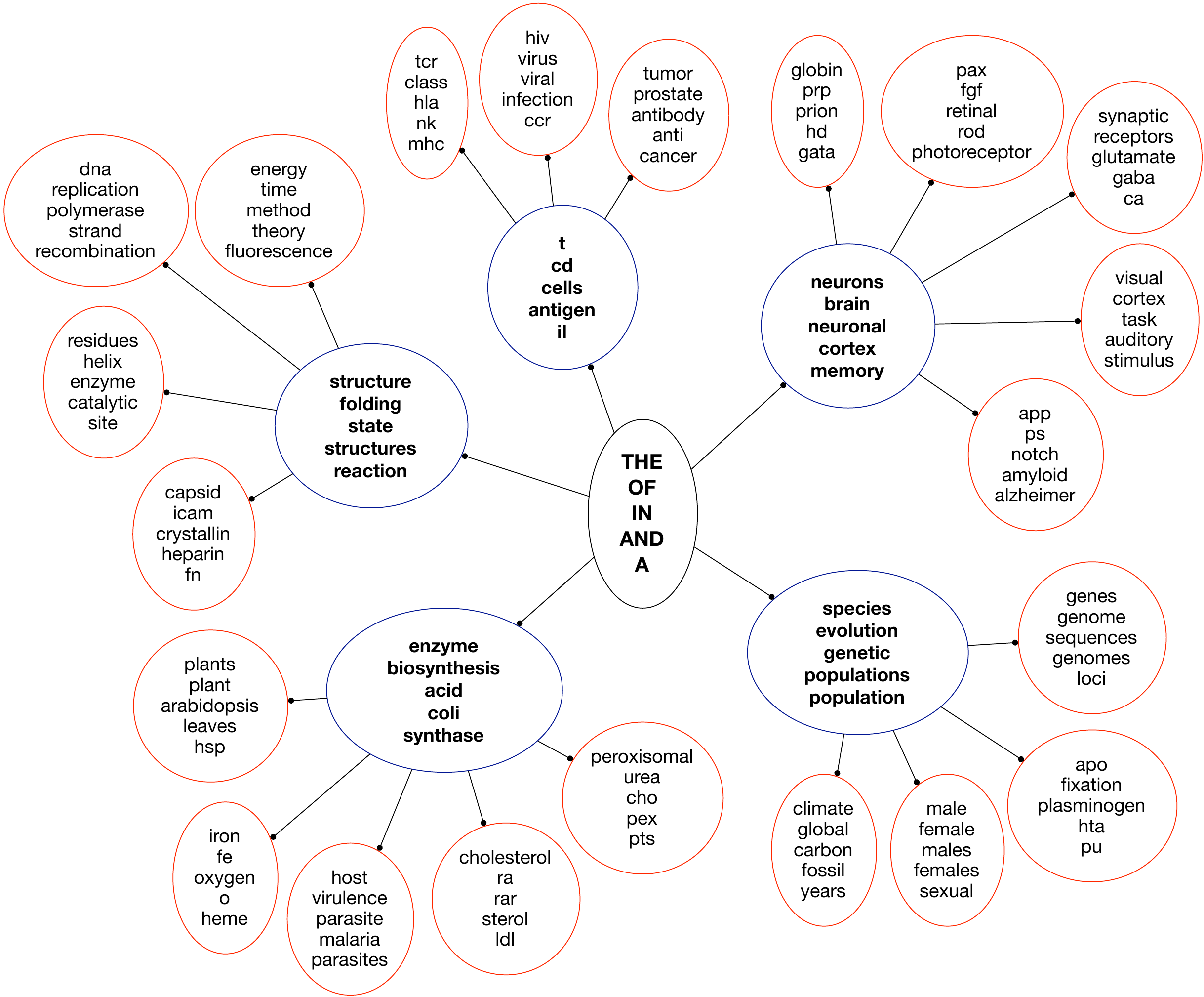}
  \end{center}
  \caption{\label{fig:PNAS} A portion of the hierarchy learned from
    the 12,913 abstracts of the \emph{Proceedings of the National
      Academy of Sciences} from 1991--2001.  The vocabulary was
    restricted to the 7,200 terms that occurred in more than five
    documents, yielding a corpus of 2.3M words.  The learned
    hierarchy, of which only a portion is illustrated, contains 56
    topics.  Note that the $\gamma$ parameter is fixed at a smaller
    value, to provide a reasonably sized topic hierarchy with the
    significantly larger corpus.}
\end{figure}

Finally, in \myfig{PNAS}, we illustrate a portion of the analysis of a
collection of 12,913 abstracts from the {\em Proceedings of the National
Academy of Sciences} from 1991 to 2001.  An underlying hierarchical structure
of the content of the journal has been discovered, dividing articles into
groups such as neuroscience, immunology, population genetics and
enzymology.

In all three of these examples, the same posterior inference algorithm
with the same hyperparameters yields very different tree structures for
different corpora.  Models of fixed tree structure force us to commit to
one in advance of seeing the data.  The nested Chinese restaurant process
at the heart of hLDA provides a flexible solution to this difficult
problem.

\subsection{Comparison to LDA}
\label{sec:quant}

In this section we present experiments comparing hLDA to its
non-hierarchical precursor, LDA.  We use the infinite-depth hLDA
model; the per-document distribution over levels is not truncated.  We
use predictive held-out likelihood to compare the two approaches
quantitatively, and we present examples of LDA topics in order to
provide a qualitative comparison of the methods.  LDA has been shown
to yield good predictive performance relative to competing unigram
language models, and it has also been argued that the topic-based
analysis provided by LDA represents a qualitative improvement on
competing language models~\citep{Blei:2003b,Griffiths:2006}.  Thus LDA
provides a natural point of comparison.

There are several issues that must be borne in mind in comparing hLDA
to LDA.  First, in LDA the number of topics is a fixed parameter, and
a model selection procedure is required to choose the number of
topics.  (A Bayesian nonparametric solution to this can be obtained
with the hierarchical Dirichlet process~\citep{Teh:2007}.)  Second,
given a set of topics, LDA places no constraints on the usage of the
topics by documents in the corpus; a document can place an arbitrary
probability distribution on the topics.  In hLDA, on the other hand, a
document can only access the topics that lie along a single path in
the tree.  In this sense, LDA is significantly more flexible than
hLDA.

This flexibility of LDA implies that for large corpora we can expect
LDA to dominate hLDA in terms of predictive performance (assuming that
the model selection problem is resolved satisfactorily and assuming
that hyperparameters are set in a manner that controls overfitting).
Thus, rather than trying to simply optimize for predictive performance
within the hLDA family and within the LDA family, we have instead
opted to first run hLDA to obtain a posterior distribution over the
number of topics, and then to conduct multiple runs of LDA for a range
of topic cardinalities bracketing the hLDA result.  This provides an
hLDA-centric assessment of the consequences (for predictive
performance) of using a hierarchy versus a flat model.

We used predictive held-out likelihood as a measure of performance.
The procedure is to divide the corpus into $D_1$ observed documents
and $D_2$ held-out documents, and approximate the conditional
probability of the held-out set given the training set
\begin{equation}
  \label{eq:BF}
  p(\vct{w}^{\textrm{held-out}}_{1}, \ldots, \vct{w}^{\textrm{held-out}}_{D_2} \g
  \vct{w}^{\textrm{obs}}_{1}, \ldots, \vct{w}^{\textrm{obs}}_{D_1}, {\cal M}),
\end{equation}
where ${\cal M}$ represents a model, either LDA or hLDA.  We employed
collapsed Gibbs sampling for both models and integrated out all the
hyperparameters with priors.  We used the same prior for those
hyperparameters that exist in both models.

To approximate this predictive quantity, we run two samplers.  First,
we collect 100 samples from the posterior distribution of latent
variables given the observed documents, taking samples 100 iterations
apart and using a burn-in of 2000 samples.  For each of these
\textit{outer samples}, we collect 800 samples of the latent variables
given the held-out documents and approximate their conditional
probability given the outer sample with the harmonic
mean~\citep{Kass:1995}.  Finally, these conditional probabilities are
averaged to obtain an approximation to \myeq{BF}.

\begin{figure}
  \begin{center}
    \includegraphics[width=0.80\textwidth]{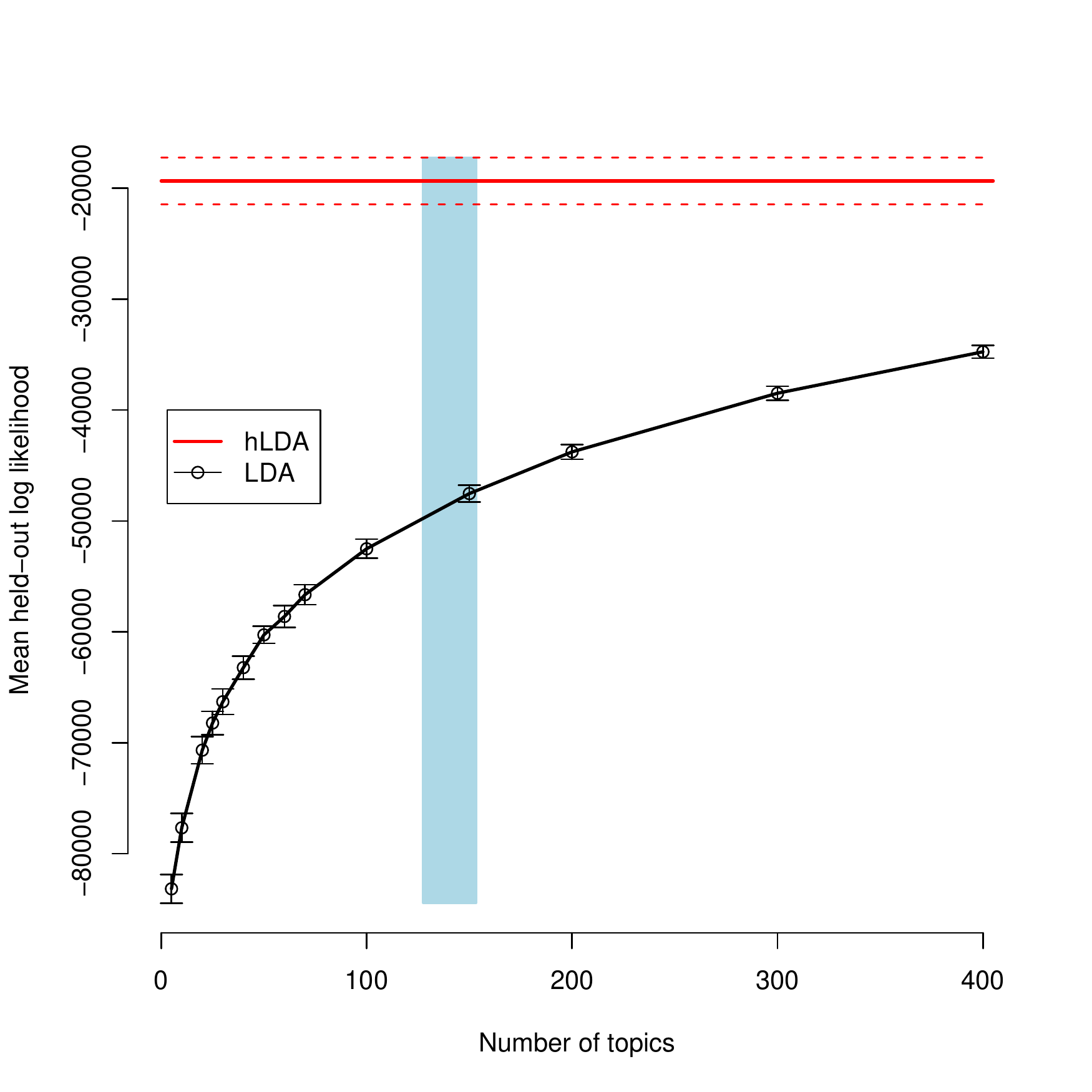}
  \end{center}
  \caption{\label{fig:lda-hlda} The held-out predictive log likelihood
    for hLDA compared to the same quantity for LDA as a function of
    the number of topics.  The shaded blue region is centered at the
    mean number of topics in the hierarchies found by hLDA (and has
    width equal to twice the standard error).}
\end{figure}

\begin{figure}
  \begin{center}
    \includegraphics[width=\textwidth]{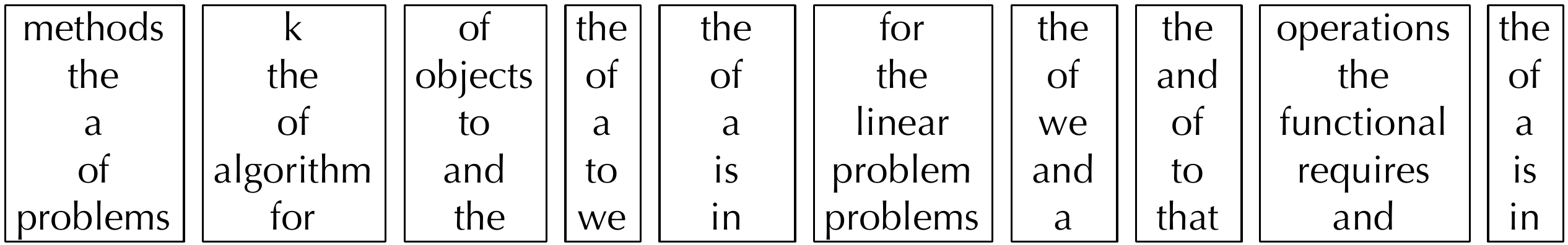}
  \end{center}
  \caption{\label{fig:lda-hlda-topics} The five most probable words
    for each of ten randomly chosen topics from an LDA model fit to
    fifty topics.}
\end{figure}

\myfig{lda-hlda} illustrates the five-fold cross-validated held-out
likelihood for hLDA and LDA on the JACM corpus.  The figure also
provides a visual indication of the mean and variance of the posterior
distribution over topic cardinality for hLDA; the mode is
approximately a hierarchy with 140 topics.  For LDA, we plot the
predictive likelihood in a range of topics around this value.

We see that at each fixed topic cardinality in this range of topics,
hLDA provides significantly better predictive performance than LDA.
As discussed above, we eventually expect LDA to dominate hLDA for
large numbers of topics.  In a large range near the hLDA mode,
however, the constraint that documents pick topics along single paths
in a hierarchy yields superior performance.  This suggests that the 
hierarchy is useful not only for interpretation, but also for capturing 
predictive statistical structure.

% (Even when the number of topics is equal to
% 400, which is not illustrated in the figure, LDA does not yet reach
% the hLDA score, though it is within the margin of error.)

To give a qualitative sense of the relative degree of interpretability
of the topics that are found using the two approaches,
\myfig{lda-hlda-topics} illustrates ten LDA topics chosen randomly
from a 50-topic model.  As these examples make clear, the LDA topics
are generally less interpretable than the hLDA topics.  In particular,
function words are given high probability throughout.  In practice, to
sidestep this issue, corpora are often stripped of function words
before fitting an LDA model.  While this is a reasonable ad-hoc
solution for (English) text, it is not a general solution that can be
used for non-text corpora, such as visual scenes.  Even more
importantly, there is no notion of abstraction in the LDA topics.  The
notion of multiple levels of abstraction requires a model such as
hLDA.

In summary, if interpretability is the goal, then there are strong
reasons to prefer hLDA to LDA.  If predictive performance is the goal,
then hLDA may well remain the preferred method if there is a
constraint that a relatively small number of topics should be used.
When there is no such constraint, LDA may be preferred.  These
comments also suggest, however, that an interesting direction for
further research is to explore the feasibility of a model that
combines the defining features of the LDA and hLDA models.  As we
described in \mysec{hlda}, it may be desirable to consider an
hLDA-like hierarchical model that allows each document to exhibit
multiple paths along the tree.  This might be appropriate for
collections of long documents, such as full-text articles, which tend
to be more heterogeneous than short abstracts.

% DMB: note comment on gamma above.  will this raise flags?

\section{Discussion}
\label{sec:discussion}

In this paper, we have shown how the nested Chinese restaurant 
process can be used to define prior distributions on recursive 
data structures.  We have also shown how this prior can be 
combined with a topic model to yield a Bayesian nonparametric 
methodology for analyzing document collections in terms of hierarchies 
of topics.  Given a collection of documents, we use MCMC sampling 
to learn an underlying thematic structure that provides a useful 
abstract representation for data visualization and summarization.

We emphasize that no knowledge of the topics of the collection or the
structure of the tree are needed to infer a hierarchy from data.  We
have demonstrated our methods on collections of abstracts from three
different scientific journals, showing that while the content of these
different domains can vary significantly, the statistical principles
behind our model make it possible to recover meaningful sets of topics
at multiple levels of abstraction, and organized in a tree.

The Bayesian nonparametric framework underlying our work makes it
possible to define probability distributions and inference procedures
over countably infinite collections of objects.  There has been other
recent work in artificial intelligence in which probability
distributions are defined on infinite objects via concepts from
first-order logic~\citep{Milch:2005,Pasula:2001,Poole:2007}.  While
providing an expressive language, this approach does not necessarily
yield structures that are amenable to efficient posterior inference.
Our approach reposes instead on combinatorial structure---the
exchangeability of the Dirichlet process as a distribution on
partitions---and this leads directly to a posterior inference
algorithm that can be applied effectively to large-scale learning
problems.

The hLDA model draws on two complementary insights---one from
statistics, the other from computer science. From statistics, we take
the idea that it is possible to work with general stochastic processes
as prior distributions, thus accommodating latent structures that vary
in complexity. This is the key idea behind Bayesian nonparametric 
methods.  In recent years, these models have been extended to include
spatial models~\citep{Duan:2006} and grouped data~\citep{Teh:2007},
and Bayesian nonparametric methods now enjoy new applications in
computer vision~\citep{Sudderth:2005},
bioinformatics~\citep{Xing:2007}, and natural language
processing~\citep{Li:2007,Teh:2007,Goldwater:2006a,Goldwater:2006,Johnson:2007,Liang:2007}.

From computer science, we take the idea that the representations we
infer from data should be richly structured, yet admit efficient
computation.  This is a growing theme in Bayesian nonparametric 
research. For example, one line of recent research has explored 
stochastic processes involving multiple binary features rather than
clusters~\citep{Griffiths:2006a,Thibaux:2007,Teh:2007a}.  A parallel
line of investigation has explored alternative posterior inference
techniques for Bayesian nonparametric models, providing more efficient
algorithms for extracting this latent structure. Specifically,
variational methods, which replace sampling with optimization, have
been developed for Dirichlet process mixtures to further increase
their applicability to large-scale data analysis
problems~\citep{Blei:2005,Kurihara:2006}.

The hierarchical topic model that we explored in this paper is just
one example of how this synthesis of statistics and computer science
can produce powerful new tools for the analysis of complex data.
However, this example showcases the two major strengths of the
Bayesian nonparametric approach. First, the use of the nested CRP
means that the model does not start with a fixed set of topics or
hypotheses about their relationship, but grows to fit the data at
hand.  Thus, we learn a topology but do not commit to it; the tree can
grow as new documents about new topics and subtopics are
observed. Second, despite the fact that this results in a very rich
hypothesis space, containing trees of arbitrary depth and branching
factor, it is still possible to perform approximate probabilistic
inference using a simple algorithm. This combination of flexible,
structured representations and efficient inference makes nonparametric
Bayesian methods uniquely promising as a formal framework for learning
with flexible data structures.

\subsection*{Acknowledgments}

We thank Edo Airoldi for providing the PNAS data, and we thank three
anonymous reviewers for their insightful comments.  David M. Blei is
supported by ONR 175-6343, NSF CAREER 0745520, and grants from Google
and Microsoft Research.  Thomas L. Griffiths is supported by NSF grant
BCS-0631518 and the DARPA CALO project.  Michael I. Jordan is
supported by grants from Google and Microsoft Research.

% hack to deal with not being able to put cites in captions.
\nocite{Goldberg:1986}
\bibliographystyle{acmtrans}
\bibliography{dp-nested-arxiv}

\begin{thebibliography}{}

\bibitem[\protect\citeauthoryear{Airoldi, Blei, Fienberg, and Xing}{Airoldi
  et~al\mbox{.}}{2008}]{Airoldi:2008}
{\sc Airoldi, E.}, {\sc Blei, D.}, {\sc Fienberg, S.}, {\sc and} {\sc Xing, E.}
  2008.
\newblock Mixed membership stochastic blockmodels.
\newblock {\em Journal of Machine Learning Research\/}~{\em 9}, 1981--2014.

\bibitem[\protect\citeauthoryear{Albert and Barabasi}{Albert and
  Barabasi}{2002}]{Albert:2001}
{\sc Albert, R.} {\sc and} {\sc Barabasi, A.} 2002.
\newblock Statistical mechanics of complex networks.
\newblock {\em Reviews of Modern Physics\/}~{\em 74,\/}~1, 47--97.

\bibitem[\protect\citeauthoryear{Aldous}{Aldous}{1985}]{Aldous:1985}
{\sc Aldous, D.} 1985.
\newblock Exchangeability and related topics.
\newblock In {\em {\'E}cole d'{E}t{\'e} de Probabilit{\'e}s de Saint-Flour,
  XIII---1983}. Springer, Berlin, Germany, 1--198.

\bibitem[\protect\citeauthoryear{Antoniak}{Antoniak}{1974}]{Antoniak:1974}
{\sc Antoniak, C.} 1974.
\newblock Mixtures of {D}irichlet processes with applications to {B}ayesian
  nonparametric problems.
\newblock {\em The Annals of Statistics\/}~{\em 2}, 1152--1174.

\bibitem[\protect\citeauthoryear{Barabasi and Reka}{Barabasi and
  Reka}{1999}]{Barabasi:1999}
{\sc Barabasi, A.} {\sc and} {\sc Reka, A.} 1999.
\newblock Emergence of scaling in random networks.
\newblock {\em Science\/}~{\em 286,\/}~5439, 509--512.

\bibitem[\protect\citeauthoryear{Bernardo and Smith}{Bernardo and
  Smith}{1994}]{Bernardo:1994}
{\sc Bernardo, J.} {\sc and} {\sc Smith, A.} 1994.
\newblock {\em {B}ayesian Theory}.
\newblock John Wiley \& Sons Ltd., Chichester, UK.

\bibitem[\protect\citeauthoryear{Billingsley}{Billingsley}{1995}]{Billingsley:%
1995}
{\sc Billingsley, P.} 1995.
\newblock {\em Probability and Measure}.
\newblock Wiley-Interscience, New York, NY.

\bibitem[\protect\citeauthoryear{Blei, Griffiths, Jordan, and Tenenbaum}{Blei
  et~al\mbox{.}}{2003}]{Blei:2003}
{\sc Blei, D.}, {\sc Griffiths, T.}, {\sc Jordan, M.}, {\sc and} {\sc
  Tenenbaum, J.} 2003.
\newblock Hierarchical topic models and the nested {C}hinese restaurant
  process.
\newblock In {\em Advances in Neural Information Processing Systems 16}. MIT
  Press, Cambridge, MA, 17--24.

\bibitem[\protect\citeauthoryear{Blei and Jordan}{Blei and
  Jordan}{2003}]{Blei:2003a}
{\sc Blei, D.} {\sc and} {\sc Jordan, M.} 2003.
\newblock Modeling annotated data.
\newblock In {\em Proceedings of the 26th annual International ACM SIGIR
  Conference on Research and Development in Information Retrieval}. ACM Press,
  127--134.

\bibitem[\protect\citeauthoryear{Blei and Jordan}{Blei and
  Jordan}{2005}]{Blei:2005}
{\sc Blei, D.} {\sc and} {\sc Jordan, M.} 2005.
\newblock Variational inference for {D}irichlet process mixtures.
\newblock {\em Journal of Bayesian Analysis\/}~{\em 1}, 121--144.

\bibitem[\protect\citeauthoryear{Blei and Lafferty}{Blei and
  Lafferty}{2006}]{Blei:2006a}
{\sc Blei, D.} {\sc and} {\sc Lafferty, J.} 2006.
\newblock Dynamic topic models.
\newblock In {\em Proceedings of the 23rd International Conference on Machine
  Learning}. ACM Press, New York, NY, 113--120.

\bibitem[\protect\citeauthoryear{Blei and Lafferty}{Blei and
  Lafferty}{2007}]{Blei:2007}
{\sc Blei, D.} {\sc and} {\sc Lafferty, J.} 2007.
\newblock A correlated topic model of \textit{Science}.
\newblock {\em Annals of Applied Statistics\/}~{\em 1}, 17--35.

\bibitem[\protect\citeauthoryear{Blei and Lafferty}{Blei and
  Lafferty}{2009}]{Blei:2009}
{\sc Blei, D.} {\sc and} {\sc Lafferty, J.} 2009.
\newblock Topic models.
\newblock In {\em Text Mining: Theory and Applications}. Taylor and Francis,
  London, UK.

\bibitem[\protect\citeauthoryear{Blei, Ng, and Jordan}{Blei
  et~al\mbox{.}}{2003}]{Blei:2003b}
{\sc Blei, D.}, {\sc Ng, A.}, {\sc and} {\sc Jordan, M.} 2003.
\newblock Latent {D}irichlet allocation.
\newblock {\em Journal of Machine Learning Research\/}~{\em 3}, 993--1022.

\bibitem[\protect\citeauthoryear{Chakrabarti, Dom, Agrawal, and
  Raghavan}{Chakrabarti et~al\mbox{.}}{1998}]{Chakrabarti:1998}
{\sc Chakrabarti, S.}, {\sc Dom, B.}, {\sc Agrawal, R.}, {\sc and} {\sc
  Raghavan, P.} 1998.
\newblock Scalable feature selection, classification and signature generation
  for organizing large text databases into hierarchical topic taxonomies.
\newblock {\em The VLDB Journal\/}~{\em 7}, 163--178.

\bibitem[\protect\citeauthoryear{Cimiano, Hotho, and Staab}{Cimiano
  et~al\mbox{.}}{2005}]{Cimiano:2005}
{\sc Cimiano, P.}, {\sc Hotho, A.}, {\sc and} {\sc Staab, S.} 2005.
\newblock Learning concept hierarchies from text corpora using formal concept
  analysis.
\newblock {\em Journal of Artificial Intelligence Research\/}~{\em 24},
  305--339.

\bibitem[\protect\citeauthoryear{Deerwester, Dumais, Landauer, Furnas, and
  Harshman}{Deerwester et~al\mbox{.}}{1990}]{Deerwester:1990}
{\sc Deerwester, S.}, {\sc Dumais, S.}, {\sc Landauer, T.}, {\sc Furnas, G.},
  {\sc and} {\sc Harshman, R.} 1990.
\newblock Indexing by latent semantic analysis.
\newblock {\em Journal of the American Society of Information Science\/}~6,
  391--407.

\bibitem[\protect\citeauthoryear{Devroye, Gy\"orfi, and Lugosi}{Devroye
  et~al\mbox{.}}{1996}]{Devroye96}
{\sc Devroye, L.}, {\sc Gy\"orfi, L.}, {\sc and} {\sc Lugosi, G.} 1996.
\newblock {\em A Probabilistic Theory of Pattern Recognition}.
\newblock Springer-Verlag, New York, NY.

\bibitem[\protect\citeauthoryear{Dietz, Bickel, and Scheffer}{Dietz
  et~al\mbox{.}}{2007}]{Dietz:2007}
{\sc Dietz, L.}, {\sc Bickel, S.}, {\sc and} {\sc Scheffer, T.} 2007.
\newblock Unsupervised prediction of citation influences.
\newblock In {\em Proceedings of the 24th International Conference on Machine
  Learning}. ACM Press, New York, NY, 233--240.

\bibitem[\protect\citeauthoryear{Drinea, Enachesu, and Mitzenmacher}{Drinea
  et~al\mbox{.}}{2006}]{Drinea:2006}
{\sc Drinea, E.}, {\sc Enachesu, M.}, {\sc and} {\sc Mitzenmacher, M.} 2006.
\newblock Variations on random graph models for the web.
\newblock Tech. Rep. TR-06-01, Harvard University.

\bibitem[\protect\citeauthoryear{Duan, Guindani, and Gelfand}{Duan
  et~al\mbox{.}}{2007}]{Duan:2006}
{\sc Duan, J.}, {\sc Guindani, M.}, {\sc and} {\sc Gelfand, A.} 2007.
\newblock Generalized spatial {D}irichlet process models.
\newblock {\em Biometrika\/}~{\em 94}, 809--825.

\bibitem[\protect\citeauthoryear{Duda, Hart, and Stork}{Duda
  et~al\mbox{.}}{2000}]{Duda:2000}
{\sc Duda, R.}, {\sc Hart, P.}, {\sc and} {\sc Stork, D.} 2000.
\newblock {\em Pattern Classification}.
\newblock Wiley-Interscience, New York, NY.

\bibitem[\protect\citeauthoryear{Dumais and Chen}{Dumais and
  Chen}{2000}]{Dumais:2000}
{\sc Dumais, S.} {\sc and} {\sc Chen, H.} 2000.
\newblock Hierarchical classification of web content.
\newblock In {\em Proceedings of the 23rd Annual International ACM SIGIR
  conference on Research and Development in Information Retrieval}. ACM Press,
  New York, NY, 256--263.

\bibitem[\protect\citeauthoryear{Escobar and West}{Escobar and
  West}{1995}]{Escobar:1995}
{\sc Escobar, M.} {\sc and} {\sc West, M.} 1995.
\newblock Bayesian density estimation and inference using mixtures.
\newblock {\em Journal of the American Statistical Association\/}~{\em 90},
  577--588.

\bibitem[\protect\citeauthoryear{Fei-Fei and Perona}{Fei-Fei and
  Perona}{2005}]{Fei-Fei:2005}
{\sc Fei-Fei, L.} {\sc and} {\sc Perona, P.} 2005.
\newblock A {B}ayesian hierarchical model for learning natural scene
  categories.
\newblock {\em IEEE Computer Vision and Pattern Recognition\/}, 524--531.

\bibitem[\protect\citeauthoryear{Ferguson}{Ferguson}{1973}]{Ferguson:1973}
{\sc Ferguson, T.} 1973.
\newblock A {B}ayesian analysis of some nonparametric problems.
\newblock {\em Annals of Statistics\/}~{\em 1}, 209--230.

\bibitem[\protect\citeauthoryear{Gelfand and Smith}{Gelfand and
  Smith}{1990}]{Gelfand:1990}
{\sc Gelfand, A.} {\sc and} {\sc Smith, A.} 1990.
\newblock Sampling based approaches to calculating marginal densities.
\newblock {\em Journal of the American Statistical Association\/}~{\em 85},
  398--409.

\bibitem[\protect\citeauthoryear{Gelman, Carlin, Stern, and Rubin}{Gelman
  et~al\mbox{.}}{1995}]{Gelman:1995}
{\sc Gelman, A.}, {\sc Carlin, J.}, {\sc Stern, H.}, {\sc and} {\sc Rubin, D.}
  1995.
\newblock {\em Bayesian Data Analysis}.
\newblock Chapman \& Hall, London, UK.

\bibitem[\protect\citeauthoryear{Geman and Geman}{Geman and
  Geman}{1984}]{Geman:1984}
{\sc Geman, S.} {\sc and} {\sc Geman, D.} 1984.
\newblock Stochastic relaxation, {G}ibbs distributions and the {B}ayesian
  restoration of images.
\newblock {\em IEEE Transactions on Pattern Analysis and Machine
  Intelligence\/}~{\em 6}, 721--741.

\bibitem[\protect\citeauthoryear{Goldberg and Tarjan}{Goldberg and
  Tarjan}{1986}]{Goldberg:1986}
{\sc Goldberg, A.} {\sc and} {\sc Tarjan, R.} 1986.
\newblock A new approach to the maximum flow problem.
\newblock {\em Journal of the Association for Computing Machinery\/}~{\em
  35,\/}~4, 921--940.

\bibitem[\protect\citeauthoryear{Goldwater, Griffiths, and Johnson}{Goldwater
  et~al\mbox{.}}{2006a}]{Goldwater:2006}
{\sc Goldwater, S.}, {\sc Griffiths, T.}, {\sc and} {\sc Johnson, M.} 2006a.
\newblock Contextual dependencies in unsupervised word segmentation.
\newblock In {\em Proceedings of the 21st International Conference on
  Computational Linguistics and 44th Annual Meeting of the Association for
  Computational Linguistics}. Association for Computational Linguistics,
  Stroudsburg, PA, 673--680.

\bibitem[\protect\citeauthoryear{Goldwater, Griffiths, and Johnson}{Goldwater
  et~al\mbox{.}}{2006b}]{Goldwater:2006a}
{\sc Goldwater, S.}, {\sc Griffiths, T.}, {\sc and} {\sc Johnson, M.} 2006b.
\newblock Interpolating between types and tokens by estimating power-law
  generators.
\newblock In {\em Advances in Neural Information Processing Systems 18}. MIT
  Press, Cambridge, MA, 459--467.

\bibitem[\protect\citeauthoryear{Griffiths and Ghahramani}{Griffiths and
  Ghahramani}{2006}]{Griffiths:2006a}
{\sc Griffiths, T.} {\sc and} {\sc Ghahramani, Z.} 2006.
\newblock Infinite latent feature models and the {I}ndian buffet process.
\newblock In {\em Advances in Neural Information Processing Systems 18}. MIT
  Press, Cambridge, MA, 475--482.

\bibitem[\protect\citeauthoryear{Griffiths and Steyvers}{Griffiths and
  Steyvers}{2004}]{Griffiths:2004a}
{\sc Griffiths, T.} {\sc and} {\sc Steyvers, M.} 2004.
\newblock Finding scientific topics.
\newblock {\em Proceedings of the National Academy of Science\/}~{\em 101},
  5228--5235.

\bibitem[\protect\citeauthoryear{Griffiths and Steyvers}{Griffiths and
  Steyvers}{2006}]{Griffiths:2006}
{\sc Griffiths, T.} {\sc and} {\sc Steyvers, M.} 2006.
\newblock Probabilistic topic models.
\newblock In {\em Latent Semantic Analysis: A Road to Meaning}, {T.~Landauer},
  {D.~McNamara}, {S.~Dennis}, {and} {W.~Kintsch}, Eds. Erlbaum, Hillsdale, NJ.

\bibitem[\protect\citeauthoryear{Hastie, Tibshirani, and Friedman}{Hastie
  et~al\mbox{.}}{2001}]{Hastie:2001}
{\sc Hastie, T.}, {\sc Tibshirani, R.}, {\sc and} {\sc Friedman, J.} 2001.
\newblock {\em The Elements of Statistical Learning}.
\newblock Springer, New York, NY.

\bibitem[\protect\citeauthoryear{Heller and Ghahramani}{Heller and
  Ghahramani}{2005}]{Heller:2005}
{\sc Heller, K.} {\sc and} {\sc Ghahramani, Z.} 2005.
\newblock Bayesian hierarchical clustering.
\newblock In {\em Proceedings of the 22nd International Conference on Machine
  Learning}. ACM Press, Cambridge, MA, 297--304.

\bibitem[\protect\citeauthoryear{Hjort, Holmes, M\"uller, and Walker}{Hjort
  et~al\mbox{.}}{2009}]{HjortEtAl}
{\sc Hjort, N.}, {\sc Holmes, C.}, {\sc M\"uller, P.}, {\sc and} {\sc Walker,
  S.} 2009.
\newblock {\em Bayesian Nonparametrics: Principles and Practice}.
\newblock Cambridge University Press, Cambridge, UK.

\bibitem[\protect\citeauthoryear{Hofmann}{Hofmann}{1999a}]{Hofmann:1999}
{\sc Hofmann, T.} 1999a.
\newblock The cluster-abstraction model: {U}nsupervised learning of topic
  hierarchies from text data.
\newblock In {\em Proceedings of the 15th International Joint Conferences on
  Artificial Intelligence}. Morgan Kaufmann, San Francisco, CA, 682--687.

\bibitem[\protect\citeauthoryear{Hofmann}{Hofmann}{1999b}]{Hofmann:1999a}
{\sc Hofmann, T.} 1999b.
\newblock Probabilistic latent semantic indexing.
\newblock In {\em Proceedings of the 22nd Annual ACM SIGIR Conference on
  Research and Development in Information Retrieval}. ACM Press, New York, NY,
  50--57.

\bibitem[\protect\citeauthoryear{Johnson, Griffiths, and S.}{Johnson
  et~al\mbox{.}}{2007}]{Johnson:2007}
{\sc Johnson, M.}, {\sc Griffiths, T.}, {\sc and} {\sc S., G.} 2007.
\newblock Adaptor grammars: {A} framework for specifying compositional
  nonparametric {B}ayesian models.
\newblock In {\em Advances in Neural Information Processing Systems 19}. MIT
  Press, Cambridge, MA, 641--648.

\bibitem[\protect\citeauthoryear{Johnson and Kotz}{Johnson and
  Kotz}{1977}]{Johnson:1977}
{\sc Johnson, N.} {\sc and} {\sc Kotz, S.} 1977.
\newblock {\em Urn Models and Their Applications: {A}n Approach to Modern
  Discrete Probability Theory}.
\newblock Wiley, New York, NY.

\bibitem[\protect\citeauthoryear{Jordan}{Jordan}{2000}]{Jordan:2000}
{\sc Jordan, M.~I.} 2000.
\newblock Graphical models.
\newblock {\em Statistical Science\/}~{\em 19}, 140--155.

\bibitem[\protect\citeauthoryear{Kass and Raftery}{Kass and
  Raftery}{1995}]{Kass:1995}
{\sc Kass, R.} {\sc and} {\sc Raftery, A.} 1995.
\newblock {B}ayes factors.
\newblock {\em Journal of the American Statistical Association\/}~{\em 90},
  773--795.

\bibitem[\protect\citeauthoryear{Koller and Sahami}{Koller and
  Sahami}{1997}]{Koller:1997}
{\sc Koller, D.} {\sc and} {\sc Sahami, M.} 1997.
\newblock Hierarchically classifying documents using very few words.
\newblock In {\em Proceedings of the 14th International Conference on Machine
  Learning}. Morgan Kaufmann, San Francisco, CA, 170--178.

\bibitem[\protect\citeauthoryear{Krapivsky and Redner}{Krapivsky and
  Redner}{2001}]{Krapivsky:2000}
{\sc Krapivsky, P.} {\sc and} {\sc Redner, S.} 2001.
\newblock Organization of growing random networks.
\newblock {\em Physical Review E\/}~{\em 63,\/}~6.

\bibitem[\protect\citeauthoryear{Kurihara, Welling, and Vlassis}{Kurihara
  et~al\mbox{.}}{2007}]{Kurihara:2006}
{\sc Kurihara, K.}, {\sc Welling, M.}, {\sc and} {\sc Vlassis, N.} 2007.
\newblock Accelerated variational {D}irichlet process mixtures.
\newblock In {\em Advances in Neural Information Processing Systems 19}. MIT
  Press, Cambridge, MA, 761--768.

\bibitem[\protect\citeauthoryear{Larsen and Aone}{Larsen and
  Aone}{1999}]{Larsen:1999}
{\sc Larsen, B.} {\sc and} {\sc Aone, C.} 1999.
\newblock Fast and effective text mining using linear-time document clustering.
\newblock In {\em Proceedings of the 5th ACM SIGKDD International Conference on
  Knowledge Discovery and Data Mining}. ACM Press, New York, NY, 16--22.

\bibitem[\protect\citeauthoryear{Lauritzen}{Lauritzen}{1996}]{Lauritzen:1996}
{\sc Lauritzen, S.~L.} 1996.
\newblock {\em Graphical Models}.
\newblock Oxford University Press, Oxford, UK.

\bibitem[\protect\citeauthoryear{Li, Blei, and McCallum}{Li
  et~al\mbox{.}}{2007}]{Li:2007}
{\sc Li, W.}, {\sc Blei, D.}, {\sc and} {\sc McCallum, A.} 2007.
\newblock Nonparametric {B}ayes pachinko allocation.
\newblock In {\em Proceedings of the 23rd Conference on Uncertainty in
  Artificial Intelligence}. AUAI Press, Menlo Park, CA.

\bibitem[\protect\citeauthoryear{Liang, Petrov, Klein, and Jordan}{Liang
  et~al\mbox{.}}{2007}]{Liang:2007}
{\sc Liang, P.}, {\sc Petrov, S.}, {\sc Klein, D.}, {\sc and} {\sc Jordan, M.}
  2007.
\newblock The infinite {PCFG} using hierarchical {D}irichlet processes.
\newblock In {\em Proceedings of the 2007 Joint Conference on Empirical Methods
  in Natural Language Processing and Computational Natural Language Learning}.
  Association for Computational Linguistics, Stroudsburg, PA, 688--697.

\bibitem[\protect\citeauthoryear{Liu}{Liu}{1994}]{Liu:1994}
{\sc Liu, J.} 1994.
\newblock The collapsed {G}ibbs sampler in {B}ayesian computations with
  application to a gene regulation problem.
\newblock {\em Journal of the American Statistical Association\/}~{\em
  89,\/}~958--966.

\bibitem[\protect\citeauthoryear{MacEachern and Muller}{MacEachern and
  Muller}{1998}]{MacEachern:1998}
{\sc MacEachern, S.} {\sc and} {\sc Muller, P.} 1998.
\newblock Estimating mixture of {D}irichlet process models.
\newblock {\em Journal of Computational and Graphical Statistics\/}~{\em 7},
  223--238.

\bibitem[\protect\citeauthoryear{Marlin}{Marlin}{2003}]{Marlin:2003}
{\sc Marlin, B.} 2003.
\newblock Modeling user rating profiles for collaborative filtering.
\newblock In {\em Advances in Neural Information Processing Systems 16}. MIT
  Press, Cambridge, MA, 627--634.

\bibitem[\protect\citeauthoryear{McCallum, Corrada-Emmanuel, and Wang}{McCallum
  et~al\mbox{.}}{2004}]{McCallum:2004}
{\sc McCallum, A.}, {\sc Corrada-Emmanuel, A.}, {\sc and} {\sc Wang, X.} 2004.
\newblock The author-recipient-topic model for topic and role discovery in
  social networks: {E}xperiments with {E}nron and academic email.
\newblock Tech. rep., University of Massachusetts, Amherst.

\bibitem[\protect\citeauthoryear{McCallum, Nigam, Rennie, and Seymore}{McCallum
  et~al\mbox{.}}{1999}]{McCallum:1999}
{\sc McCallum, A.}, {\sc Nigam, K.}, {\sc Rennie, J.}, {\sc and} {\sc Seymore,
  K.} 1999.
\newblock Building domain-specific search engines with machine learning
  techniques.
\newblock In {\em Proceedings of the AAAI Spring Symposium on Intelligent
  Agents in Cyberspace}. AAAI Press, Menlo Park, CA.

\bibitem[\protect\citeauthoryear{Milch, Marthi, Sontag, Ong, and Kobolov}{Milch
  et~al\mbox{.}}{2005}]{Milch:2005}
{\sc Milch, B.}, {\sc Marthi, B.}, {\sc Sontag, D.}, {\sc Ong, D.}, {\sc and}
  {\sc Kobolov, A.} 2005.
\newblock Approximate inference for infinite contingent {B}ayesian networks.
\newblock In {\em Proceedings of 10th International Workshop on Artificial
  Intelligence and Statistics}. The Society for Artificial Intelligence and
  Statistics, NJ.

\bibitem[\protect\citeauthoryear{Mimno and McCallum}{Mimno and
  McCallum}{2007}]{Mimno:2007}
{\sc Mimno, D.} {\sc and} {\sc McCallum, A.} 2007.
\newblock Organizing the {OCA}: {L}earning faceted subjects from a library of
  digital books.
\newblock In {\em Proceedings of the 7th ACM/IEEE-CS Joint Conference on
  Digital libraries}. ACM Press, New York, NY, 376--385.

\bibitem[\protect\citeauthoryear{Neal}{Neal}{2000}]{Neal:2000}
{\sc Neal, R.} 2000.
\newblock {M}arkov chain sampling methods for {D}irichlet process mixture
  models.
\newblock {\em Journal of Computational and Graphical Statistics\/}~{\em 9},
  249--265.

\bibitem[\protect\citeauthoryear{Newman, Chemudugunta, and Smyth}{Newman
  et~al\mbox{.}}{2006}]{Newman:2007}
{\sc Newman, D.}, {\sc Chemudugunta, C.}, {\sc and} {\sc Smyth, P.} 2006.
\newblock Statistical entity-topic models.
\newblock In {\em Proceedings of the 12th ACM SIGKDD International Conference
  on Knowledge Discovery and Data Mining}. ACM Press, New York, NY, 680--686.

\bibitem[\protect\citeauthoryear{Pasula and Russell}{Pasula and
  Russell}{2001}]{Pasula:2001}
{\sc Pasula, H.} {\sc and} {\sc Russell, S.} 2001.
\newblock Approximate inference for first-order probabilistic languages.
\newblock In {\em Proceedings of the 17th International Joint Conferences on
  Artificial Intelligence}. Morgan Kaufmann, San Francisco, CA, 741--748.

\bibitem[\protect\citeauthoryear{Pitman}{Pitman}{2002}]{Pitman:2002}
{\sc Pitman, J.} 2002.
\newblock {\em Combinatorial Stochastic Processes}.
\newblock Lecture Notes for St. Flour Summer School. Springer-Verlag, New York,
  NY.

\bibitem[\protect\citeauthoryear{Poole}{Poole}{2007}]{Poole:2007}
{\sc Poole, D.} 2007.
\newblock Logical generative models for probabilistic reasoning about
  existence, roles and identity.
\newblock In {\em Proceedings of the 22nd AAAI Conference on Artificial
  Intelligence}. AAAI Press, Menlo Park, CA, 1271--1279.

\bibitem[\protect\citeauthoryear{Pritchard, Stephens, and Donnelly}{Pritchard
  et~al\mbox{.}}{2000}]{Pritchard:2000}
{\sc Pritchard, J.}, {\sc Stephens, M.}, {\sc and} {\sc Donnelly, P.} 2000.
\newblock Inference of population structure using multilocus genotype data.
\newblock {\em Genetics\/}~{\em 155}, 945--959.

\bibitem[\protect\citeauthoryear{Robert and Casella}{Robert and
  Casella}{2004}]{Robert:2004}
{\sc Robert, C.} {\sc and} {\sc Casella, G.} 2004.
\newblock {\em Monte {C}arlo Statistical Methods}.
\newblock Springer-Verlag, New York, NY.

\bibitem[\protect\citeauthoryear{Rosen-Zvi, Griffiths, Steyvers, and
  Smith}{Rosen-Zvi et~al\mbox{.}}{2004}]{Rosen-Zvi:2004}
{\sc Rosen-Zvi, M.}, {\sc Griffiths, T.}, {\sc Steyvers, M.}, {\sc and} {\sc
  Smith, P.} 2004.
\newblock The author-topic model for authors and documents.
\newblock In {\em Proceedings of the 20th Conference on Uncertainty in
  Artificial Intelligence}. AUAI Press, Menlo Park, CA, 487--494.

\bibitem[\protect\citeauthoryear{Sanderson and Croft}{Sanderson and
  Croft}{1999}]{Sanderson:1999}
{\sc Sanderson, M.} {\sc and} {\sc Croft, B.} 1999.
\newblock Deriving concept hierarchies from text.
\newblock In {\em Proceedings of the 22nd Annual International ACM SIGIR
  Conference on Research and Development in Information Retrieval}. ACM, New
  York, NY, 206--213.

\bibitem[\protect\citeauthoryear{Sethuraman}{Sethuraman}{1994}]{Sethuraman:199%
4}
{\sc Sethuraman, J.} 1994.
\newblock A constructive definition of {D}irichlet priors.
\newblock {\em Statistica Sinica\/}~{\em 4}, 639--650.

\bibitem[\protect\citeauthoryear{Stoica and Hearst}{Stoica and
  Hearst}{2004}]{Stoica:2004}
{\sc Stoica, E.} {\sc and} {\sc Hearst, M.} 2004.
\newblock Nearly-automated metadata hierarchy creation.
\newblock In {\em Companion Proceedings of HLT-NAACL}. Boston, MA.

\bibitem[\protect\citeauthoryear{Sudderth, Torralba, Freeman, and
  Willsky}{Sudderth et~al\mbox{.}}{2005}]{Sudderth:2005}
{\sc Sudderth, E.}, {\sc Torralba, A.}, {\sc Freeman, W.}, {\sc and} {\sc
  Willsky, A.} 2005.
\newblock Describing visual scenes using transformed {D}irichlet processes.
\newblock In {\em Advances in Neural Information Processing Systems 18}. MIT
  Press, Cambridge, MA, 1297--1306.

\bibitem[\protect\citeauthoryear{Teh, Gorur, and Ghahramani}{Teh
  et~al\mbox{.}}{2007}]{Teh:2007a}
{\sc Teh, Y.}, {\sc Gorur, D.}, {\sc and} {\sc Ghahramani, Z.} 2007.
\newblock Stick-breaking construction for the {I}ndian buffet process.
\newblock In {\em Proceedings of 11th International Workshop on Artificial
  Intelligence and Statistics}. The Society for Artificial Intelligence and
  Statistics, NJ.

\bibitem[\protect\citeauthoryear{Teh, Jordan, Beal, and Blei}{Teh
  et~al\mbox{.}}{2007}]{Teh:2007}
{\sc Teh, Y.}, {\sc Jordan, M.}, {\sc Beal, M.}, {\sc and} {\sc Blei, D.} 2007.
\newblock Hierarchical {D}irichlet processes.
\newblock {\em Journal of the American Statistical Association\/}~{\em 101},
  1566--1581.

\bibitem[\protect\citeauthoryear{Thibaux and Jordan}{Thibaux and
  Jordan}{2007}]{Thibaux:2007}
{\sc Thibaux, R.} {\sc and} {\sc Jordan, M.} 2007.
\newblock Hierarchical beta processes and the {I}ndian buffet process.
\newblock In {\em Proceedings of 11th International Workshop on Artificial
  Intelligence and Statistics}. The Society for Artificial Intelligence and
  Statistics, NJ.

\bibitem[\protect\citeauthoryear{Titterington, Smith, and Makov}{Titterington
  et~al\mbox{.}}{1985}]{Titterington:1985}
{\sc Titterington, D.}, {\sc Smith, A.}, {\sc and} {\sc Makov, E.} 1985.
\newblock {\em Statistical Analysis of Finite Mixture Distributions}.
\newblock Wiley, Chichester, UK.

\bibitem[\protect\citeauthoryear{Vaithyanathan and Dom}{Vaithyanathan and
  Dom}{2000}]{Vaithyanathan:2000}
{\sc Vaithyanathan, S.} {\sc and} {\sc Dom, B.} 2000.
\newblock Model-based hierarchical clustering.
\newblock In {\em Proceedings of the 16th Conference on Uncertainty in
  Artificial Intelligence}. Morgan Kaufmann, San Francisco, CA, 599--608.

\bibitem[\protect\citeauthoryear{Xing, Jordan, and Sharan}{Xing
  et~al\mbox{.}}{2007}]{Xing:2007}
{\sc Xing, E.}, {\sc Jordan, M.}, {\sc and} {\sc Sharan, R.} 2007.
\newblock Bayesian haplotype inference via the {D}irichlet process.
\newblock {\em Journal of Computational Biology\/}~{\em 14}, 267--284.

\bibitem[\protect\citeauthoryear{Zamir and Etzioni}{Zamir and
  Etzioni}{1998}]{Zamir:1998}
{\sc Zamir, O.} {\sc and} {\sc Etzioni, O.} 1998.
\newblock Web document clustering: {A} feasibility demonstration.
\newblock In {\em Proceedings of the 21st Annual International ACM SIGIR
  Conference on Research and Development in Information Retrieval}. ACM Press,
  New York, NY, 46--54.

\end{thebibliography}

\end{document}